# Introducing HALC: A general pipeline for the systematic and reliable construction of prompts for automated coding with LLMs in the computational social sciences

Andreas Reich
andreas.reich@uni-hohenheim.de
https://orcid.org/0000-0002-2426-6490

Claudia Thoms
claudia.thoms@uni-hohenheim.de
https://orcid.org/0000-0002-6601-1170

Tobias Schrimpf
tobias.schrimpf@uni-hohenheim.de
https://orcid.org/0009-0001-6257-5948
@SchrimpfTobias

**Funding**

The authors received no financial support for the research, authorship, and/or publication of this article.

**Disclosure statement & Disclosure of interest**

The authors report there are no competing interests to declare.

**Data availability statement & data disposition**

The dataset and scripts created for this paper can be found here:

https://osf.io/f24d7/?view_only=5cd1ded18ca5496e9d6fd5a5d6703230

**Acknowledgements**

We would like to thank Tarek Bahiani, Sophie Mann, and Sabine Sonnemann for their assistance with the content analysis.

*Use of generative AI tools*

For this paper we employed the following AI tools. For language improvement the tool *DeepL* was mainly used. The LLMs *ChatGPT* (versions o-3, 4o, 4.1) and *Gemini* (versions 2.5 – 3.1 Pro) as well as *QWQ-32B* were used for programming assistance. Some of these LLMs were also used for language improvements.

**Title**

Introducing HALC: A general pipeline for the systematic and reliable construction of prompts for automated coding with LLMs in the computational social sciences

**Abstract**

LLMs are seeing widespread use for task automation, including automated coding in the social sciences. However, even though researchers have proposed different prompting strategies, their effectiveness varies across LLMs and tasks. Often trial and error practices are still widespread. Our study aims to fill this gap and evaluate how LLMs can be used in a systematic and transparent way to produce reliable codings in content analyses. We propose HALC—a general pipeline that allows for the systematic and reliable construction of prompts for any given coding task and model. We develop this pipeline based on current literature and findings of a prestudy investigating consistency and influencing factors of LLM codings. We also apply HALC on two other datasets covering different thematic contexts, document types, languages, and coding units to test its applicability. Based on more than three million LLM requests, our results demonstrate that the pipeline is capable of identifying prompts for reliable codings in different settings. We also discuss shortcomings and further potential for development.

## Introduction

The rise of transformer-based (Vaswani et al., 2017) language models and especially the widespread use of generative large language models (LLMs) (Ouyang et al., 2022) has prompted a plethora of research concerning their effective use. LLMs harbor great potential and especially open-source LLMs promise to democratize access to and broaden possibilities for automated coding. However, reliable automated coding with LLMs presents several challenges. Finding adequate prompting strategies to utilize LLMs effectively is still demanding, especially since there are many different options that can be employed, and various models may behave differently even when using the same strategies. Hence, the open-endedness that is the greatest strength of LLMs also presents the biggest challenge to automated coding, which must yield reliable, reproducible, and comprehensible results. This has led researchers to seek suitable frameworks for identifying adequate prompts. Despite these challenges, the rise of generative LLMs presents an opportunity to enhance and streamline automated quantitative content analyses by addressing the scalability and adaptability limitations of traditional approaches.

Comparing different approaches, humans are usually the central bottleneck (see Figure 1). Although manual content analyses are quite flexible in terms of the type and complexity of the recorded content, they are limited by the available resources. Classic automated text classification using supervised machine learning (ML) is one possible solution. However, it usually requires extensive and high-quality training material to achieve good results (Gilardi et al., 2023). This also makes it very specialized, often rendering models trained in such ways unusable in other contexts (Kroon et al., 2024).

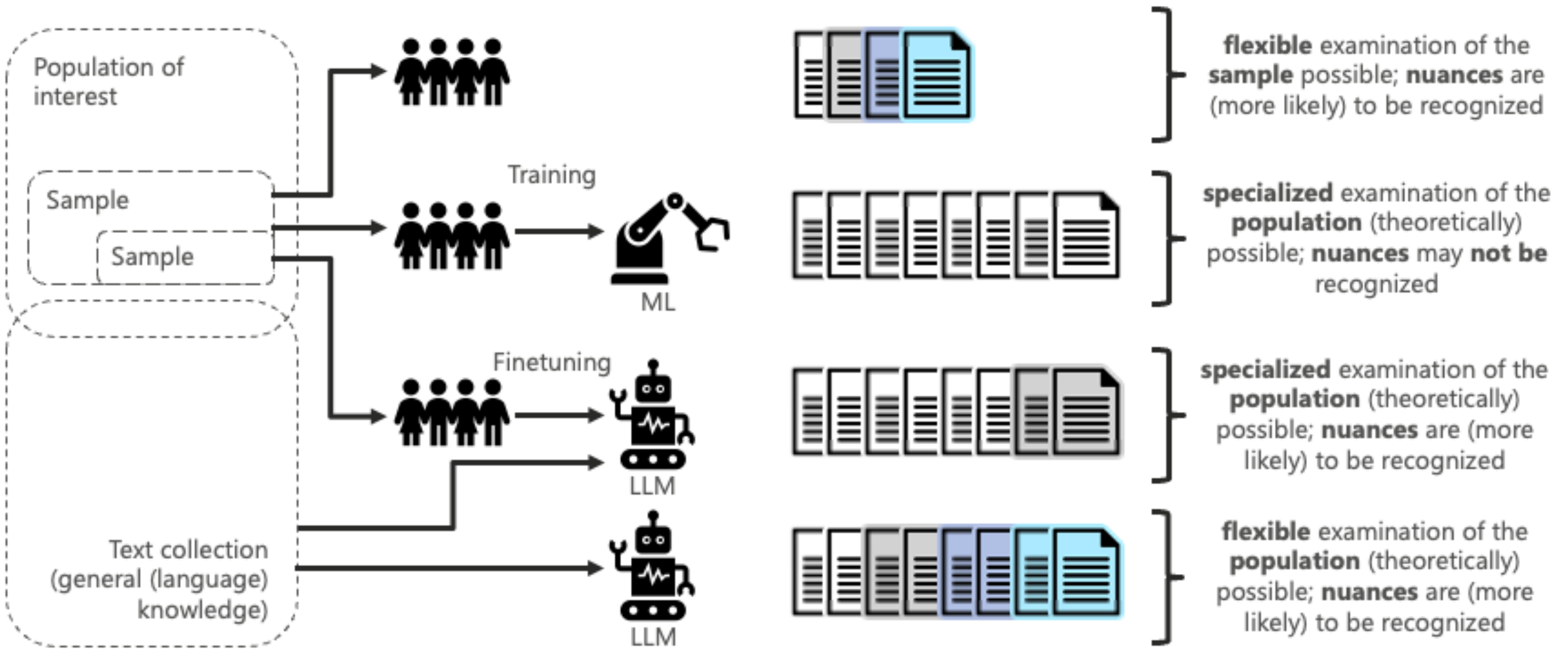

**Figure 1.** Comparison of different methodological approaches for content analyses

Generative LLMs promise to enhance quantitative content analyses with their capability to process and create text. These potentials are already being explored in the social sciences (e.g., Gilardi et al., 2023; Pilny et al., 2024). LLMs bring general and linguistic knowledge from processing large amounts of text, which can be built upon (Laurer et al., 2024). Context-dependent meanings are better captured, allowing for more nuanced analyses than, for example, dictionary-based approaches (Kroon et al., 2024). Since LLMs are pretrained on general knowledge (Radford et al., 2018), they require less training data for fine-tuning or can even be used directly in zero-shot settings (Pilny et al., 2024). In this way, zero-shot classification offers a substantial degree of flexibility and generalizability (Brown et al., 2020), making it particularly interesting for automated content analyses. Furthermore, the integration of human and machine coding could benefit from the fact that LLMs can be controlled in a relatively natural way by so-called prompts (Zamfirescu-Pereira et al., 2023). The transfer of coding instructions for human coding to LLMs could therefore be easier to implement than with classic machine learning, which is far more technical (Törnberg, 2024a).

However, the specialized knowledge and unpredictable reactions of LLMs, due to their practically non-deterministic operation, make them difficult to work with. Despite initial proposals for its design (e.g., Chew et al., 2023), the systematic combination of manual and LLM-based content analyses has been challenging. Developing adequate prompts for querying LLMs is an important step in this process: It is not only important to consider the specifics of LLMs (e.g., in terms of language, formatting, and space), but also to address the question of how human and machine coding can best build upon each other to ensure alignment (Ouyang et al., 2022). This is not only relevant for efficiency, but also for validity and reliability.

Authors such as Törnberg (2024a) have argued that “human coders should not be considered ground truth” (p. 73), since interpretations made by LLMs could be superior to those made by human coders. This has led to the recommendation of a mutual adaptation of human and machine coders (Törnberg, 2024a) or a “codebook co-development” (Chew et al., 2023), i.e. the integration of the LLM during the operationalization phase to take its understanding into account. Critically, this can be countered with two questions: How is this LLM superiority determined, if not by human reasoning? And what is more important, validity and completeness or codability and reliability? Although we also see the merits of these perspectives, we are more in line with Haim et al. (2023) who have stated that “human coders are still one of the cornerstones of content analysis in the tradition of communication science. Their coding can make *a*, if not *the*, difference to the quality of any content analysis” (p. 284). They are “essential as an external validity criterion” (Niemann-Lenz et al., 2023, p. 347). Therefore, in our view, human codings are the starting point for every computational extension of content analytical research.

Our study aims to demonstrate how LLMs can be utilized in a systematic and transparent way to produce reliable codings in content analyses. We begin with reviewing current literature about challenges, influencing factors, and existing approaches to combining LLMs and human coders. Next, we investigate important prerequisites for systematic and reliable usage of LLMs like coding consistency and reliability in a prestudy. We then combine our findings and insights from literature into a pipeline. The pipeline is based on existing methodology used in manual content analyses (e.g., Früh, 2017) in the social sciences and on prompt engineering from the field of computer science (e.g., Korzynski et al., 2023; Lo, 2023). By leveraging few expert codings as ground truth, we anchor the evaluation of LLM outputs in established scientific standards, ensuring validity and reliability. This approach not only honors the foundational role of human coders (as emphasized by Haim et al., 2023), but also opens pathways for scalable and adaptable automated coding. Furthermore, our approach promises to reduce technical hurdles and systematize the process of combining manual and automated content analyses that make use of LLMs. Finally, we explore the pipeline’s applicability using additional datasets in our main study. We conclude by reflecting on our results and potential future applications, showing that systematic prompting can improve the usefulness and reliability of LLMs for automated coding.

## Literature review

### Challenges related to the use of LLMs for automated content analyses

Research on LLMs and their potential as a tool in the social sciences has virtually exploded with the release of *OpenAI*'s *ChatGPT* in November 2022 (Breuer, 2023). However, the breathtaking speed at which developments have progressed also presents several methodological challenges. First, new models and prompting strategies are continually emerging. This makes the question concerning the generalizability of findings based on one model to other models even more relevant (e.g., Weber & Reichardt, 2023). After all, there are many factors involved in their performance so that differences between models cannot be automatically ruled out (Alizadeh et al., 2024). Second, the accessibility, apparent ease of use, and impressive performance of LLMs in text processing and production are offset by a relatively opaque mode of operation. This is particularly true of commercial LLMs (Breuer, 2023). But even open-source models retain a certain black-box character due to their mode of operation, as LLMs are at least difficult to interpret or understand (Sudmann, 2020) and are practically non-deterministic systems. This raises concerns, particularly regarding the validity and reliability of LLMs as tools for content analysis or text annotation (Belz et al., 2021). Third, these conditions favor a publication culture that relies heavily on grey literature and preprints. The result is a massive increase in the volume of scholarly output, making it harder to keep track of the work being done. It is no wonder, then, that "LLM-based text annotation has become something of an academic Wild West" (Törnberg, 2024a, p. 68), where neither researchers nor reviewers can rely on established standards to ensure the quality of research. The first challenge is one of several reasons why continued research on the use of LLMs in empirical research is warranted, even though the rapid development of the field will continue to limit the generalizability of findings, including our own. With respect to the second and third challenge, however, the literature provides several conceptual and practical insights. In the following, we review these contributions as a basis for our aim of providing researchers with recommendations and procedures to systematically build on manual content analyses using LLMs.

## Factors affecting validity and reliability

Research[1] shows that different factors affect the validity and reliability of model responses. Similar considerations apply to human coding. According to Niemann-Lenz et al. (2023), the characteristics of the coders, the categories, and the coding process influence coding behavior and results. For coders, sociodemographic characteristics, personality traits, coding experience, and motivation are influential. Regarding category types, the distinction between manifest and latent constructs plays a decisive role. In the coding process, the material, the codebook, and the coder's training are important. In our view, a similar distinction can be made for LLM-based coding. Here, model-specific properties correspond to coder characteristics, the nature of the coded constructs remains a property of the categories, and prompt design and task setup parallel aspects of the coding process. We therefore use this threefold distinction to guide our selection of factors, rather than to provide an exhaustive list of all possible

[1] Of course, the challenges described above also influence our work and our literature review. We therefore try to provide targeted insights into relevant studies that address our questions, although these insights are by no means complete.

influences on LLM behavior, and focus on the following factors that are prominent in the emerging literature:

**Model size and architecture.** Generally, given similar training conditions, larger models with more parameters tend to perform better than smaller models since they can generalize complex tasks better and memorize more information. Also, different model architectures (Liu et al., 2024), attention mechanisms (Gu & Dao, 2024), and practices such as reasoning, test-time compute, or latent reasoning (Geiping et al., 2025) greatly affect model performance.

**Model settings.** LLM parameters such as temperature, seed, or top-k sampling can affect the quality of the results. Of these, temperature, which controls the randomness of the model (Ferraris et al., 2025; Törnberg, 2024a), has been shown to be important for the reliability of LLM coding. Indeed, the “creativity” (Ferraris et al., 2025) that results from this randomness has some advantages, such as potentially increasing output novelty, although often at the cost of coherence (Peeperkorn et al., 2024). In general, a lower temperature (i.e., less randomness) is associated with more consistent annotations (Gilardi et al., 2023; Reiss, 2023).

**Initial data.** When human coding is used as ground truth—and this is usually the case—the quality of this initial data has been identified as an important success factor for the quality of LLM coding, specifically its validity and reliability. In this context, two main challenges arise that affect manual and automated coding alike. First, the distribution of the coded categories matters. As is well known from human-coded reliability studies, small samples and highly skewed variables can distort reliability assessment, for example by leading to low chance-corrected reliability coefficients even when percentage agreement is high (Lacy et al., 2015; Vogelgesang & Scharkow, 2012). This equally affects the evaluation of LLM-based coding when it is benchmarked against such human ground truth. Second, meaning multiplicity in texts can lead to low reliability scores even when disagreements reflect valid disagreement rather than measurement error (Baden et al., 2023). Thus, both distributional issues and valid disagreement limit how clearly human-coded data can serve as a benchmark for LLM performance. In the context of automated content analysis, this is closely related to current debates about whether and in what sense LLMs might outperform human coders (Törnberg, 2024a). Even under these constraints, however, the quality of the human coding used as ground truth remains crucial. Data coded by experts (Pangakis et al., 2023) or determined by majority decision from multiple coders (Törnberg, 2024b) have been recommended and tend to be of higher quality. Codebooks related to such data could therefore be considered a good starting point for the development of instructions to guide LLMs. Although research has shown that a reliable codebook for human annotation does not necessarily guarantee reliable LLM coding (Reiss, 2023), it still seems clear that unreliable data is not a good ground truth to work with. This has also been shown for more basic approaches such as supervised machine learning, where manually coded data is used for model training, but it also remains relevant in the context of LLMs (Oschatz et al., 2023). Ultimately, poor quality data simply does not provide a clear basis for deciding whether model performance is good enough and should therefore be considered as a potential degradation factor.

**Type of variables coded.** Closely related to the quality of the initial data is the importance of variables of varying difficulty for coding quality. The resulting consequences for the reliability of coding are also well known in human-coded content analysis. As described by Riffe et al. (2024), more manifest concepts are easier to code reliably than latent ones; they are also less ambiguous, so their interpretation is more transferable to outsiders who are not involved in the systematic coding process, but who encounter corresponding content in everyday life. If humans already struggle with some types of variables, it is not surprising that performance differences have also been observed in LLM coding (Pangakis et al., 2023; Weber & Reichardt, 2023).

**Type of model inquiry.** What has already been established for human coding can also be established for LLMs: offsetting answers leads to better results. The so-called "wisdom of the crowd" effect (Törnberg, 2024b) can be simulated by repeated requests to an LLM to exploit the possibility of obtaining multiple plausible answers to a complex problem (Wang et al., 2023). Determining the majority decision from multiple requests, known as self-consistency prompting, is therefore an important strategy for exploiting the creativity or randomness of the model while ensuring the stability of the results (Pangakis et al., 2023; Reiss, 2023; Than et al., 2025; Wang et al., 2023).

**Prompt engineering.** Prompt engineering can be defined as "the process of constructing queries or inputs (i.e. prompts) for AI language models so as to elicit the most precise, coherent, and pertinent responses" (Lo, 2023, p. 1). Rather than making technical changes to a model, prompt engineering focuses on refining the natural language instructions given to an LLM to achieve a specific result (Ferraris et al., 2025; Törnberg, 2024a). This task has proven to be more complex than it may seem at first glance. Often it is hard to obtain good results without intensive trial and error, even among experts, as no truly established workflows have yet emerged in this process (Zamfirescu-Pereira et al., 2023). This is further complicated by the fact that the smallest changes to a prompt can have large effects—in non-deterministic settings, results have been found to vary even when identical prompts are used repeatedly (Reiss, 2023).

In sum, these and other factors, many of which involve multiple decision points, can introduce substantial variability into the process. Compared to other, more basic decisions and conditions addressed here, interacting with LLMs via prompts during the coding process stands out as a particularly flexible and thus critical source of variability, especially when prompts are optimized through intensive but irreproducible trial and error. This highlights the need for a systematic approach that harnesses the model's randomness while ensuring reproducibility.

## Human coders and LLM collaboration frameworks: some general drawbacks

Some initial attempts have been made to introduce frameworks based on the adaptation of known procedures and standards to this brave new world. Such frameworks are important to ensure that human and automated coding can build on each other in the best possible way. This is a necessary condition to avoid arbitrariness in the research process and to ensure efficiency, validity, and reliability.

Focusing on quantitative content analysis[2], authors like Chew et al. (2023), Fan et al. (2024), Pangakis et al. (2023), and Törnberg (2024a), just to name a few, have proposed frameworks to integrate human and machine coding. All these frameworks involve several steps. Basically, they build on existing codebooks and data from manual content analyses. These are used to test and, if necessary, optimize the instructions for LLMs in a conjoint feedback loop with an LLM until the result is good enough to let the LLM code the material at hand.

Although these proposals are good starting points for systematically linking human and LLM coding, there are some drawbacks. First, the co-development of the codebook, as suggested by Törnberg (2024a) and Chew et al. (2023), i.e., taking LLMs into account when developing the codebook to be used by the human coders, may prioritize codability and reliability over the validity and completeness of the operationalizations of the concepts of interest. This is not necessarily problematic and may be a fair decision. For example, it prevents the emergence of a conceptual gap between the human codebook and the LLM codebook (Pangakis et al., 2023). However, it is also a decision that can, in extreme cases, favor a more superficial analysis that does less justice to the complexity of the matter. Törnberg (2024a) himself points out this danger when he emphasizes that "it is important to come in with an explicitly articulated idea of the concept you are trying to capture, to avoid being overly influenced by the interpretations of the LLM" (p. 72).

Second, the iterative process of optimizing prompts remains opaque and inefficient. As mentioned above, minimal changes to prompts can have a large impact, and the practically non-deterministic nature of LLMs can lead to variable results from one request to the next. The iterative approach hence opens the door to a process that can drag on. In addition, without documentation, it may be unclear at the end what exactly led to the improvement and why (a point also raised by Fan et al., 2024), and whether these adjustments can be generalized to a variable- or model-independent recommendation.

## Considerations for instructing LLMs based on prompt engineering and manual content analysis

Bearing in mind the goal of systematically linking human and LLM coding, the literature points to central content-related and formal prompt characteristics that can inform a more systematic construction of prompts. Moreover, parallels between prompt engineering and classical content analysis allow us to derive guidelines that reflect requirements from both domains and potentially facilitate the translation of human coding instructions into LLM prompts.

Concerning content-related characteristics, Korzynski et al. (2023) have described four essential components of effective prompts: context description, instruction, data to be processed, and output format. Apart from the data to be processed, which is separated from the codebook during human coding, there is overlap with the components of category descriptions for human coding, namely definitions, instructions, examples, and codes or characteristics to be coded (Brosius et al., 2016; Chew et al., 2023; Früh, 2017; Neuendorf, 2002; Rössler, 2017). Context refers to specific information about the role the model is to play in the task, or more broadly, the content or thematic background. This is comparable to references regarding the goal of coding a particular category in human coding. The instruction clarifies exactly what the task is. In human coding, this corresponds to the definition of the category, which in addition to real or invented examples may include lists of indicators that clarify which terms or content can be used to identify more abstract concepts (Früh, 2017). The output format corresponds to the coding instructions and to the characteristics that can be coded for a category. In the case of LLMs, this description may also include technical aspects, e.g., if a specific data format (e.g., JSON or CSV) is explicitly requested to facilitate further processing.

[2] There is also research on the use of LLMs for qualitative content analysis (e.g., Smirnov, 2025).

Consequently, category descriptions for human coding are a good basis for prompts (e.g., Xiao et al., 2023) because they are usually based on similar components. However, prompt engineering has introduced a range of prompting strategies that can improve LLM performance. The large and constantly growing number of such strategies makes it difficult to maintain an overview. Given the dynamic nature of this field, systematic compilations of prompting strategies are likely to become outdated quickly, yet works such as those by Debnath et al. (2025) and Sahoo et al. (2025) remain helpful. Focusing on a few recurring strategies that stood out in our literature review, it becomes clear that different prompting strategies can be related to different components of the prompt. Regarding the context description, role prompting, i.e., assigning the LLM a role suited for a task, and more generally describing the background of the task can "orient the model with any necessary background information" (Törnberg, 2024a, p. 74). In terms of implementation, the only difference between the two approaches is that in the first case, also referred to as persona pattern (White et al., 2023), the LLM should take the perspective of a thematically appropriate actor (e.g., through the phrase *You are an economic analyst*), whereas in the second case the thematic context itself is named (e.g., economic news published by companies). Apart from generally describing the task at hand (Lambert et al., 2023), important strategies related to the instructional component include Chain-of-Thought prompting. This strategy prompts LLMs to take intermediate reasoning steps to perform complex reasoning (Wei et al., 2022). In a more general version, this may consist of giving the LLM an explicit list of steps to follow to complete the task (Chew et al., 2023; Lambert et al., 2023). For example, in the context of content analysis, an LLM might be asked to first carefully read the category description, then carefully read the text to be categorized, and finally assign the most appropriate code from a list of possible codes. Closely related, but not identical, are instructional prompting strategies that differ in whether a complex task is solved in a single holistic decision or via an explicitly decomposed procedure. In the strict sense, decomposed prompting means splitting a task into several separate model calls (Khot et al., 2023). In a simpler, single-prompt variant, the overall decision can be broken down into several attribute-wise judgments, for example by asking the LLM to code a set of indicators that jointly inform an overall classification. A simplified version of Chain-of-Thought prompting is Zero-Shot Chain-of-Thought prompting (Kojima et al., 2022). By adding a general phrase such as *Let's think step by step*, LLM results can improve. Although it tends to be less effective than the more extensive Chain-of-Thought prompting, it is much more space-saving, universal (i.e. topic-independent), and quicker to use. Justifications can make LLM decisions more transparent (White et al., 2023). If they are understood as the verbalization of reasoning steps, as is the case in Chain-of-Thought prompting, they could also improve the quality of decisions (Chew et al., 2023). Thus, asking the LLM to explain or give a reason for its decision may be helpful, even if the justification is not interesting per se. Finally, predefined output formats are crucial for efficiently processing LLM-generated data, with JSON being the most widely used format (Lu et al., 2025). Importantly, the chosen prompting strategies directly shape the resulting JSON structure. This is particularly evident in multi-label classification settings, where one can either request a single holistic decision or several attribute-wise decisions, each option leading to a different JSON representation (Langner et al., 2025). Similarly, requiring justifications further modifies the structure and complexity of the resulting JSON output. Results on the effects of such strategies have sometimes varied from study to study and from model to model (Weber & Reichardt, 2023). Lambert et al. (2023), for example, reported results that confirm "anecdotal observations that role prompting is less effective or not effective at all for newer models" (p. 12). This is especially true for reasoning LLMs like *deepseek-r1* (Liu et al., 2024).

Concerning the recommendations for the formal design of prompts, Lo's (2023) *CLEAR* framework for prompt engineering provides a helpful starting point for identifying key requirements. Some of these are reflected in the requirements for category systems for human coding, like completeness, discriminatory power, detail, comprehensibility, unambiguity, and precision (Brosius et al., 2016; Früh, 2017; Neuendorf, 2002; Rössler, 2017), but some are contradictory. Accordingly, prompts should be concise, logical, explicit, adaptive, and reflective. (1) *Conciseness*—which includes not only clarity but also brevity and precision—is seen as an important prerequisite for targeted guidance of LLMs. Superfluous information can confuse models and thus worsen results. This is somewhat at odds with the need for completeness and detail in human coding. On the other hand, it could also be argued that human coders could be distracted from the ultimate goal of a category by too much detail, so it might be more effective to focus on all the *important* details. (2) A *logical* prompt is structured and coherent, so that sequences or connections become clear. (3) *Explicitness* concerns unambiguous information about the content to be considered and the results to be produced, which also applies to the output format already mentioned. The last two points, (4) *adaptation* and (5) *reflection*, concern the flexible and iterative handling of prompts and the results they produce.

## Guidelines for the rule-based translation of the codebook for the systematic construction of prompts

Drawing on these descriptions of prompt engineering and its similarities with manual content analysis, some general guidelines can be established for the systematic construction of prompts based on existing codebooks. The first three points are primarily content-related, while the last three are mainly formal in nature:

**Completeness.** Identify definitions (what should be coded?), instructions (how should it be coded?), examples (how can you recognize what should be coded?), and codes (what should the output look like?) in the category description developed for the manual content analysis, and translate them into the LLM-specific context description, instructions, and output format. Add any components that are missing for use with the LLM. Select a set of prompting strategies that support these components.

**Clarity.** Use real or invented examples or, if necessary due to space constraints, limit yourself directly to indicators that clarify which expressions or content can be used to identify more abstract concepts.

**Structure.** Make sure that the components build on each other logically. Increase the coherence of your category description by using unambiguous references and transitions between sentences.

**Conciseness.** Summarize each of the components as briefly as possible. To do this, avoid filler words and repetitions, except where the latter increase coherence (see also *Structure*).

**Comprehensibility.** Use simple sentences and consistent wording when the same meaning is intended. Avoid negations. Instead of writing what is *not* meant or should *not* be coded, write what *is* meant or *should* be coded.

**Explicitness.** Formulate unambiguously and avoid imprecise formulations and expressions that introduce ambiguity into the descriptions. This also includes empty phrases (see also *Conciseness*).

Taken together, the identified content-related and formal characteristics and the derived guidelines offer a conceptual basis for systematically linking human and LLM coding through codebook-based prompts. Yet, they also highlight several unresolved issues. First, the multitude of available prompting strategies raises the question of how researchers should select and combine them in practice when aiming to align human and LLM coding. Second, it remains unclear how far researchers should go in adapting and refining codebooks and their translation into prompts, as these researcher degrees of freedom affect the transparency of LLM classifications. These questions motivate our research agenda, which we outline in the following section.

## Research agenda

We recognize that LLMs offer great potential for automated content analyses. However, we note that there are still unsettled questions about the exact behavior of LLMs in content analyses. Therefore, we first want to investigate how LLMs behave in text-based content analysis as part of a preliminary study. For example, since LLMs incorporate random processes like the temperature parameter, there have been concerns about the consistency of LLM codings. Consistency is an important prerequisite for a reproducible, reliable application in scientific research. To investigate this, we ask

*RQ1: How consistent are the results of different prompts for repeated requests to an LLM?*

Furthermore, it is uncertain which factors influence the quality of LLM coding as measured by the reliability compared to the manually coded data. Previous studies have mainly focused on prompting strategies and their influence on coding quality. However, there are various other factors that need to be considered. For example, there has been evidence that the quality of the manual coding, the calculation of the LLM coding decision (for example, by majority decision in the case of repeated coding), and also the category that is to be coded can have an influence on reliability. Furthermore, when different prompting strategies are combined (prompt permutations), the effects of their interactions must be investigated to understand the difference between better and less effective prompts. We therefore ask

*RQ2: What effects do (1) the quality of the manually coded data, (2) repeated coding with majority decision, (3) different categories (4), and different prompting strategies have on the reliability of LLM coding?*

Based on these findings and our literature review, we develop a pipeline for automated coding using LLMs. This pipeline ensures reproducible and reliable coding based on a systematic process for identifying and evaluating prompts. Finally, in our main study, we apply this pipeline to new data and models to gain initial insights into its applicability. In doing so, we focus on the reliability of the coding. To this end, we ask

*RQ3: Does applying the pipeline lead to reliable coding across different datasets, variables, languages, and LLMs?*

## Prestudy: Testing coding consistency and influencing factors for LLM codings

### Method

*Data and reliability*

We used a dataset consisting of German articles and corresponding user comments on the climate movements *Fridays for Future* and *Last Generation*. We scraped our data from three German news sites (*TAZ*, *Zeit*, and *Welt*), resulting in a total sample of 3,485 articles and 122,321 comments. From a cleaned, sampled, and manually coded dataset of 1,949 comments, which was used for a different project (Thoms et al., in press), we drew 100 random comments for the analysis of LLM coding. The nominal categories used in the current analysis were also part of that project and measured if the comments thematized the topic of climate change ($v_{climate}$) and if at least one of two climate movements ($v_{movement}$) was thematized. The first category can be considered more difficult, as the climate topic can be very subtle and less manifest in the text than, for example, the specific mention of the names of climate movements. Three trained coders conducted the manual coding process. They yielded satisfactory reliability results (see

Table 1; details concerning the reliability test can be found in Thoms et al., in press). We refer to this dataset as $N_{coders}$. For these codings, $v_{climate}$ occurred in 31% of the 100 comments and $v_{movement}$ in 48%.

To compare the effect of coders with different levels of expertise, a second dataset was created. For this, the three authors of this paper coded the same 100 random comments to obtain an expert dataset as an alternative ground truth for the LLM evaluation. Unlike $N_{coders}$, the expert coded dataset was created by coding together and negotiating the answer in cases where the experts disagreed. In other words, it was essentially created by majority decision. Choi et al. (2026) found that such team-based coding is a valid and beneficial method for establishing high-quality ground truth data. Although it was not possible to calculate the reliability between the experts as a result of this procedure, it was possible to compare the experts with the trained coders. This revealed some differences in the reliability of the two variables (see Table 1). The expert coded dataset is referred to as $N_{experts}$. For these codings, $v_{climate}$ occurred in 29% of the 100 comments and $v_{movement}$ in 53%.

**Table 1.** Reliability of the manually coded categories

| Variable | Holsti | Lotus | Std. Lotus | α |
|---|---|---|---|---|
| *Comparison of the coders* | | | | |
| Thematizing climate change ($v_{climate}$) | .82 | .91 | .82 | .61 |
| Thematizing climate movement ($v_{movement}$) | .83 | .91 | .83 | .66 |
| *Comparison of the coders and the experts* | | | | |
| Thematizing climate change ($v_{climate}$) | .76 | .88 | .76 | .43 |
| Thematizing climate movement ($v_{movement}$) | .83 | .92 | .83 | .66 |

We used a selection of common evaluation metrics to answer our questions. *Accuracy* represents the proportion of correctly classified documents relative to the total number of cases. While easy to interpret, accuracy can be misleading in skewed datasets, where one class dominates, since predicting the majority class already boosts the results (Jeni et al., 2013). To address this, we also report precision, recall, and the F1 score. *Precision* quantifies the likelihood that a document labeled as relevant is indeed correctly classified, thus reflecting how well the model minimizes false positives. *Recall* measures the proportion of truly relevant documents that are correctly identified, emphasizing the model's ability to capture all relevant cases, even at the risk of including false positives.

The *F1 Score*, the harmonic mean of precision and recall, provides a balanced measure of model performance. It is particularly valuable in imbalanced datasets (Jeni et al., 2013). In our study, we mainly report *Krippendorff's Alpha (α)*. Since it measures inter-coder reliability adjusted for chance agreement (Krippendorff, 1970), it is the most rigorous of the presented evaluation metrics.

*Technical infrastructure*

For our LLM tests in the prestudy, we used a single GPU (*NVIDIA A100 40GB*) running multiple model instances in parallel using *ollama* (2023), each instance running in a *Docker* container and orchestrated via a custom centralized queuing system. We used the default settings for the model parameters such as temperature, context length, and top_p.

Over the course of 11 months, we performed numerous analyses, first using *Mistral AI's* open-source model *Mistral 7B*[3] (Jiang et al., 2023), and later, once available, its successor *Mistral NeMo*[4]. The latter offers a larger context window, which is especially important for analyzing larger amounts of text or longer prompts while also performing better across several benchmarks like *MMLU-PRO IFEval* (Hugging Face, 2025). In total, we sent more than two million requests to our local language models for our prestudy. All requests instructed the LLMs to produce JSON output so that the results could automatically be evaluated. In addition to the two million requests used in our studies, a significant number of additional requests occurred due to failed or repeated attempts, which were common during development. These often resulted from malformed or invalid JSON output generated by the LLMs that could not be automatically repaired. Furthermore, additional requests were made during the development and testing of the automated JSON evaluation.

*Operationalization of prompting strategies*

For our study, we chose a selection of promising prompting strategies from our literature review presented before. These are shown in Table 2 together with the variants that we used for testing. The only strategy that does not stem directly from the literature is *coding elements*. It is based on the idea that human coders who are familiar with the entire codebook implicitly consider dependent variables while coding a given variable, because they keep its derivations and follow-up questions in mind. In category-based coding with an LLM, however, the model does not know the entire codebook. It therefore makes a difference whether, for example, the prompt for detecting the occurrence of a variable also refers to its subsequent evaluation, using this evaluation as an additional indicator of the variable's occurrence. The key issue of this strategy is how much information about the category of interest is made explicit in the prompt. In our tests, we translated these strategies into meaningful combinations (prompt permutations) to test different variants for their influence on the quality of LLM coding. To create the actual prompts, we combined the corresponding prompt permutations with the category descriptions from the codebook (see chapter *Guidelines for the rule-based translation of the codebook for the systematic construction of prompts*). We also included a part for the desired output format of the LLM (in our case a JSON with Boolean coding and, depending on the prompting strategy, a preceding string), and ended with the actual comment to be coded in one request. An example of such a prompt, including the original parts of the codebook that were translated to create it, can be found in the Appendix A1.

[3] https://ollama.com/library/mistral
[4] https://ollama.com/library/mistral-nemo

**Table 2.** Prompt components, their related prompting strategies and variants for implementation

| Prompt components and related prompting strategies | Description | Variants |
|---|---|---|
| *Context description* | | |
| Role prompting | Assigning a thematically appropriate role to the LLM. | (0) none<br>(1) scientist<br>(2) chatbot |
| Context information | Giving information about the background of the task. | (0) none<br>(1) with general description of the analysis context |
| *Instruction* | | |
| Task specification | Describing the task for the LLM more specifically. | (0) none<br>(1) with general description of the coding task |
| Coding strategy | Variation in the level of detail of the coding of the LLM. | (1) overall decision on the category<br>(2) detailing indicators for the category<br>(3) detailing indicators for the category with limitations[a] |
| Coding elements | Using only the original category definitions vs. enriching them with information from dependent build-up categories. | (0) none<br>(1) considering build-up elements |
| Zero-Shot Chain-of-Thought | Asking the LLM for intermediate reasoning steps. | (0) none<br>(1) "proceed step by step"<br>(2) "think step by step" |
| Chain-of-Thought | Providing the LLM with analysis steps for intermediate reasoning. | (0) none<br>(1) explanation of the analysis steps to be carried out |
| Justification | Asking the LLM for a justification of the decision. | (0) none<br>(1) with normal justification<br>(2) with detailed justification |

*Note.* [a] $v_{movement}$ has this option compared to $v_{climate}$ because of differences in the conception of the categories. The definition of $v_{climate}$ already contains a number of indicators that can be coded in detail even if the build-up elements of following categories are not considered. This is not the case for $v_{movement}$. The category itself only asks whether a movement is mentioned. A detailed coding of the indicators is only possible if the build-up elements are considered. In summary, variants 3 and 1 for $v_{movement}$ are similar, but since the wording is different, we have kept them as separate variants.

## Results

*LLM coding consistency*

Understanding prompt consistency is a prerequisite for understanding how reliable automated coding can be, because if individual prompt quality varies due to random processes within the LLM, individual prompting strategies cannot be reliably evaluated. Thus, we conducted several tests to answer our RQ1:

> *How consistent are the results of different prompts for repeated requests to an LLM?*

First, we constructed a total of 864 individual prompts based on the strategies presented in Table 2. These prompts were then used to query *Mistral 7B* and *Mistral NeMo* to code the thematization of climate change in 100 random comments. Each comment was coded once with each prompt, without repetition. We then evaluated the accuracy of the automated LLM codings by comparing them to the data coded by the trained coders ($N_{coders}$) to investigate prompt potency. Figure 2 depicts the performance of all 864 prompts after a single iteration. It is noteworthy that most prompts grouped around 75% accuracy, while some prompts performed poorly. The distributions were comparable in both LLM models. However, this analysis was only the basis for the main test of prompt consistency.

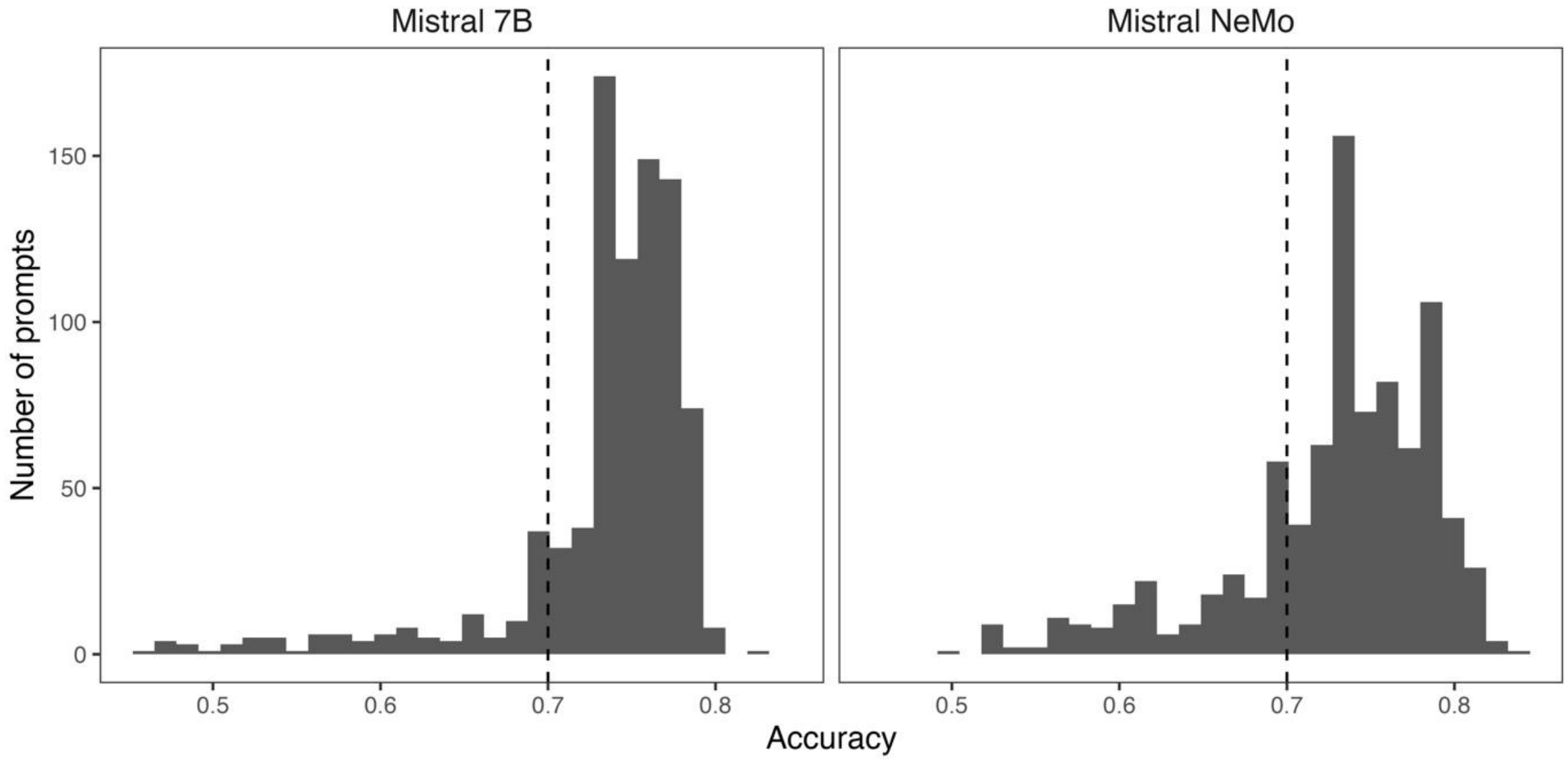

**Figure 2.** Distribution of the accuracy of the 864 prompt permutations for *Mistral 7B* and *Mistral NeMo*

The next step was to test the consistency of the prompts by coding each of the 100 comments from $N_{coders}$ 50 times. However, since iterating over all 864 prompts for *Mistral*

*7B* and *Mistral Nemo* would have meant 8.64 million requests, we decided to draw 100 random prompts to save resources. Furthermore, we only drew prompts with an accuracy of ≥ .70 to avoid investigating prompts that performed poorly. This resulted in 500,000 requests each for *Mistral 7B* and *Mistral NeMo* and gave us valuable insight into the consistency of prompts when models were asked to code the same comment repeatedly. We found similar behavior across all metrics and both models. Figure 3 depicts the range of the variance of each prompt's performance on the evaluation metrics, depending on the number of repetitions considered. For example, if a prompt achieved an accuracy of .69 in one run and .80 in the second run of coding the 100 comments, the mean would be .74, the standard deviation .08, and the variance .01. We did not determine majority decisions based on the repetitions in this step.

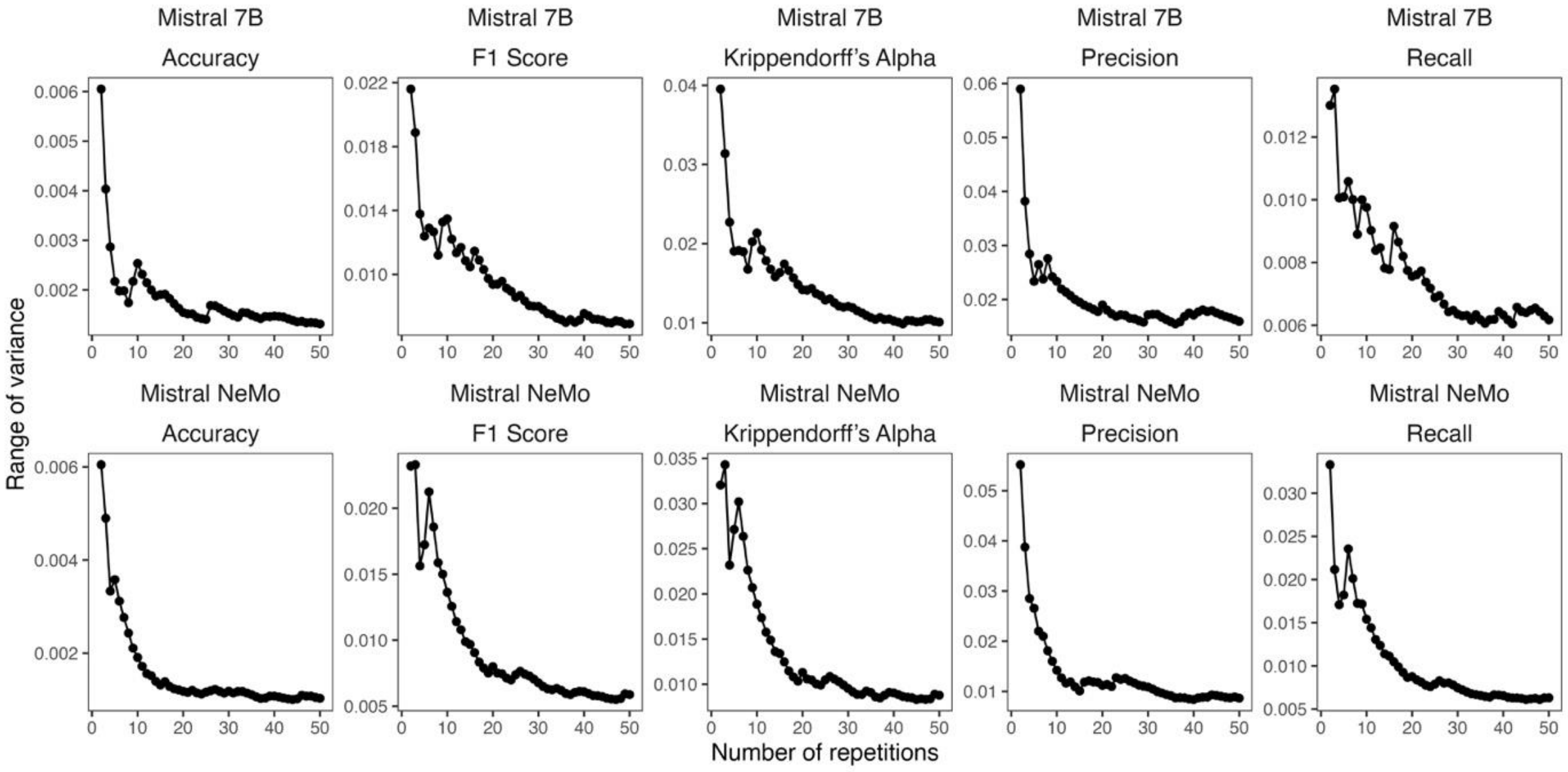

**Figure 3.** Change in range of variance of key evaluation metrics with two to 50 repetitions of 100 prompt permutations for *Mistral 7B* and *Mistral NeMo*

We identified sharp bends in the curve for 5, 15, and 25 repetitions, all of which seemed to be good candidates for coding repetitions. As we averaged the evaluation metrics of multiple requests, our estimates became more reliable. The range of variance decreased. However, while the initial gains were large, additional repetitions seemed to yield only small improvements. Looking at the scale of the range of variance across the metrics, it was clear that while repetitions were beneficial, few repetitions seemed to be sufficient. At the same time, we observed that the better prompts tended to be the more consistent ones (Figure 4). By aggregating the 50 repetitions of each prompt's coding of the 100 comments, we saw linear relationships between the mean and variance of the 100 prompts for Accuracy, F1, and Krippendorff's Alpha. Again, the results were comparable for both LLM models.

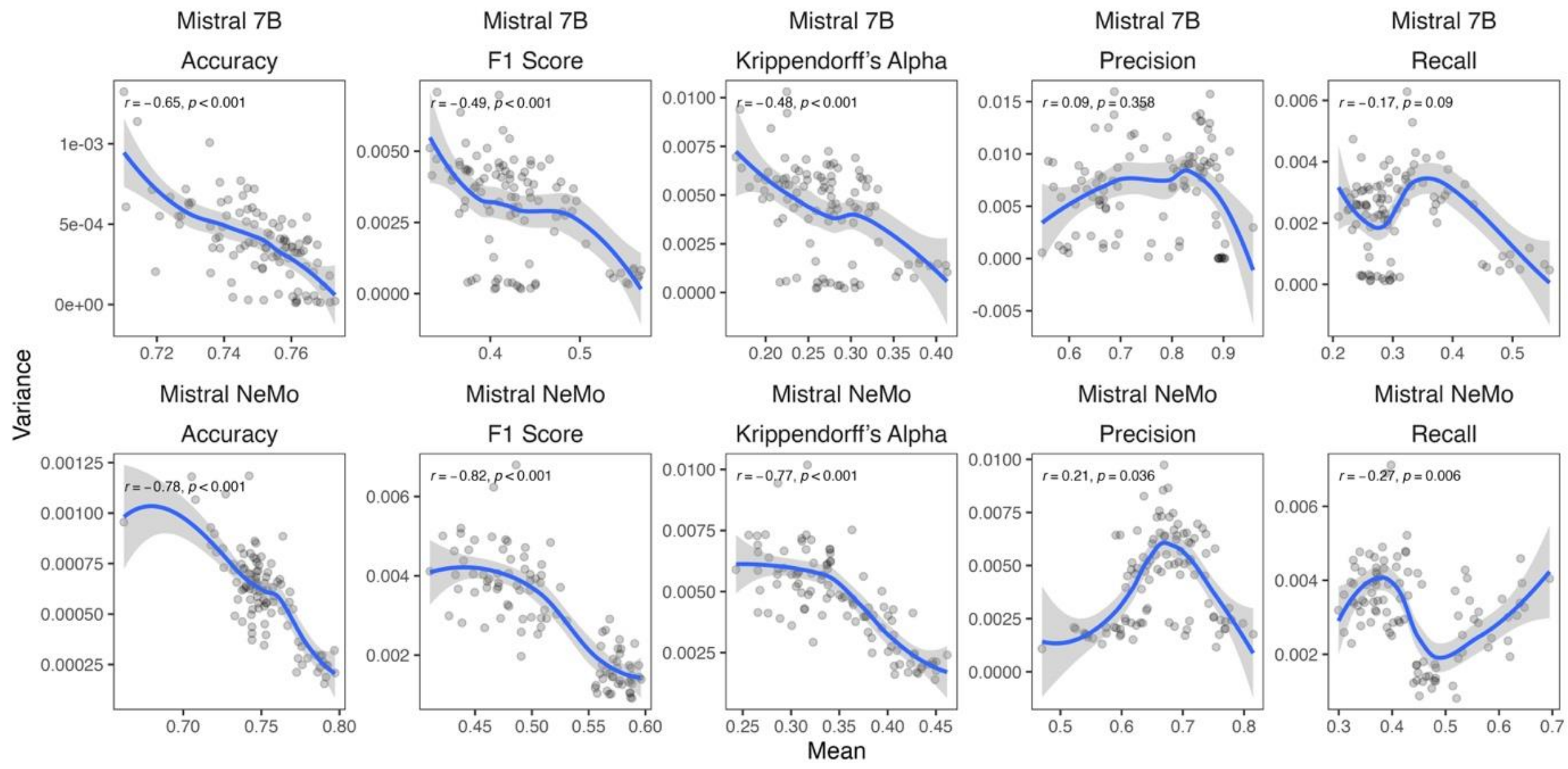

**Figure 4.** Correlation between mean and variance of key evaluation metrics with 50 repetitions of 100 prompt permutations for *Mistral 7B* and *Mistral NeMo*

Several lessons can be drawn from these results. First, repeating automated codings with the same prompts stabilized the results for all metrics studied. Five coding repetitions already led to robust evaluation metrics that provided a better basis for decisions.[5] Second, prompts with higher performance had less variance. This means that better prompts were also coded in a more stable and reliable way. These results were comparable for both models tested.

*LLM coding quality*

Based on the results for RQ1, our literature review, and technical considerations, we made several decisions for our next steps. First, we decided to continue using only the newer and better *Mistral NeMo*. Second, although RQ1 focused on the variability of LLM requests rather than their quality, it was still evident that the quality of the results was not good enough to work with. The literature points to the quality of ground truth as a possible factor influencing the validity and reliability of LLM coding. To test whether this is the case, we compared the results when expert coded data ($N_{experts}$) is used as ground truth instead of data produced by trained coders ($N_{coders}$), which was the basis of RQ1. Third, in terms of variance, repetitions proved to be important in order to make a more reliable statement about the quality of a prompt. However, the main advantage of such repetitions is the ability to determine a majority decision from multiple LLM requests to obtain better results—the so-called self-consistency prompting. According to our initial findings, five repetitions proved to be a good compromise between computational cost and stabilization of LLM results. We explicitly tested the extent to which self-consistency prompting affects response quality by comparing results that used self-consistency prompting based on five repetitions with those that did not. Fourth, the difficulty of a variable can also affect the quality of coding. This is true for both human coders and LLMs. We therefore compared the performance of two variables that we considered to be different in terms of conceptual and operational difficulty: thematizing climate change ($v_{climate}$) as a more difficult variable and thematizing the climate movement ($v_{movement}$) as an easier variable. The reliability of the human coding also indicated the different levels of difficulty of the variables (see Table 1). Finally, we investigated the impact of different prompting strategies on the results, with the aim of identifying promising strategies and combinations of strategies to achieve high reliability of LLMs with human codings and thus high validity of the results.

[5] In the case of five repetitions, the correlations between mean and variance of the key evaluation metrics are similar to those in the case of 50 repetitions (see Figure 4), but less pronounced (see Table A1 in the Appendix). Better prompts continue to be more stable.

For $v_{climate}$, the 864 prompt permutations were retained. Using the same basic prompting strategies, 648 meaningful prompt permutations (see note Table 2) were created and tested for $v_{movement}$. Due to the influencing factors tested, each of these permutations exists in multiple variants. As a result, our dataset consists of 6,048 observations. Figure 5 gives an overview of the composition of the dataset.

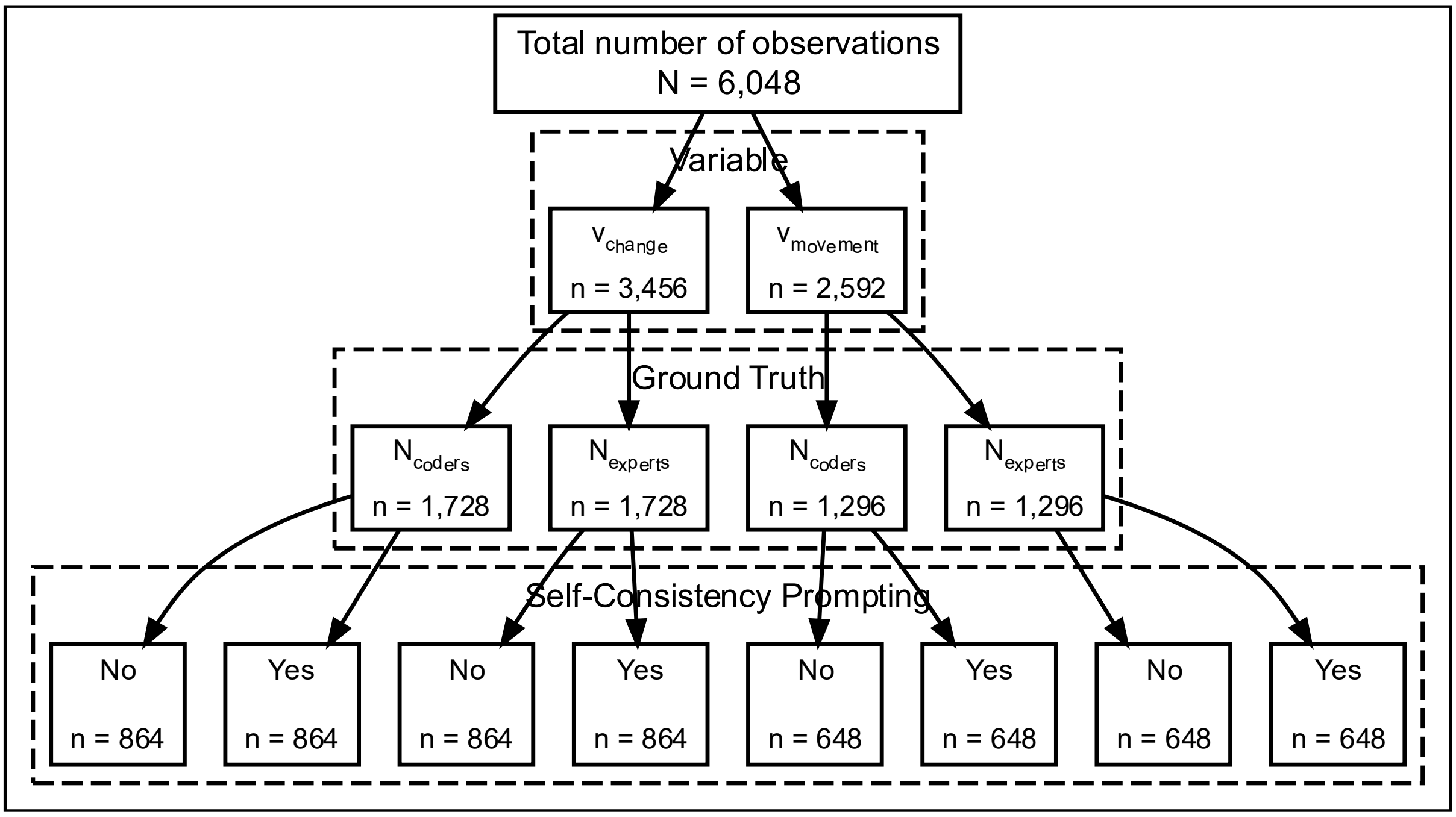

**Figure 5.** Composition of the dataset

This dataset was used to answer RQ2:

*What effects do (1) the quality of the manually coded data, (2) repeated coding with majority decision, (3) different categories (4), and different prompting strategies have on the reliability of LLM coding?*

To this end, we began our analysis with a series of multilevel regression models that tested how the different influencing factors affect Krippendorff's Alpha.[6] Our primary goal was not to predict absolute reliability levels, but to assess the relative impact of different factor groups on reliability. The need to resort to multilevel regression models resulted from the data structure. Since the prompt permutations formed the base and these permutations showed quality differences, our further tests were influenced by the specific combination of prompting strategies that make up a permutation. To account for this, we used the specific prompt permutations as random effects into which the different data variants were grouped. The influencing factors were included as fixed effects. Table 3 shows the results of the series of nested models in which the (groups of) factors are added successively. It starts with the more basic decisions and conditions in the process, such as the quality of the ground truth, the use of repetitions, and the type of variable to code, and ends with the design of the prompt itself through prompting strategies. It is noticeable that all four factor groups had significant influence on reliability and that the explained variance of the model increased with an increasing number of predictors. The positive effect of expert coding as ground truth remained significant in every model, as did the coding based on majority decision. The category to be coded also played a role, but contrary to our expectations, the less abstract thematization of the climate movement variable was less reliable than the thematization of climate change.

While it is interesting to see that not all prompting strategies seemed to have a significant effect on Krippendorff's Alpha, Model 4 in Table 3 was not ideal for analyzing them, because their effects are averaged across data variants of varying quality. To address this, we continued our analysis using the expert codings as ground truth and self-consistency prompting as the LLM coding strategy, retaining the two variables of differing complexity to obtain a more general understanding of how prompting strategies work. This dataset consisted of 1,512 prompts. Moreover, it is difficult to measure the influence of individual prompting strategies and their interactions that arise from permutations. We therefore approached the question through several steps. In a first step, we compared better and worse prompts to identify what characterized the better prompts. The criterion used for this was quite liberal: acceptable prompts were defined as those that achieved a Krippendorff's Alpha ≥ .67 (without rounding). In this regard, we followed Krippendorff's (2004, p. 241) policies, according to which our chosen threshold corresponds to the minimum acceptable value, while a Krippendorff's Alpha ≥ .80 is recommended to rely on variables. 41 of the 1,512 prompts were deemed acceptable with regard to our chosen threshold. Their Krippendorff's Alpha values were between .67 and .78. When compared to the remaining 1,471 prompts, there were some striking differences (see Table A2 in the Appendix). The two groups of prompts differed significantly in terms of coding strategy, coding elements, Chain-of-Thought prompting, and use of justifications. 40 out of 41 acceptable prompts had a detailed coding strategy and considered the build-up elements in the codebook for coding the category of interest. 34 of 41 prompts used Chain-of-Thought prompting. All but one of the acceptable prompts used a justification that was either normal length or detailed. For all other prompting strategies, there were no significant differences between acceptable and unacceptable prompts.

[6] Although Krippendorff's Alpha is theoretically bounded between -1 and 1, the observed values in our data ranged only from -0.24 to 0.78, i.e., far from the theoretical limits. Thus, the practical impact of the scale boundaries is limited in our case. In addition, diagnostic analyses (linearity, homoscedasticity, influential observations, multicollinearity, and normality of residuals and random effects) indicated no substantial violations of model assumptions. As a robustness check, we linearly rescaled Krippendorff's Alpha to the (0,1) interval and re-estimated the models using beta regression; the substantive conclusions remained unchanged. Given the large sample size and the large number of clusters, the multilevel (linear mixed-effects) regression approach was considered appropriate and robust. A comparison of different metrics for the final model is provided in the Appendix (see Figure A1).

This was telling in terms of the importance of each strategy. However, the question remained as to the extent to which certain combinations of prompting strategies were important, since taking only the significant prompting strategies into account did not necessarily lead to the best performing prompts in our dataset. It was clear that prompting strategies that occurred almost always among the acceptable prompts must also occur frequently together. However, it was less clear whether and how the remaining prompting strategies were related to this. We approached this question in a second step with the help of network logic.

Our network is based on connections between prompting strategies (nodes) and their frequency of co-occurrence (edges) regarding the dataset of acceptable reliability. In this logic, prompting strategies that occur together more often have stronger connections. Therefore, it is also possible to indirectly see how these pairwise occurrences are related to other strategies. As a result, nodes with stronger attraction are closer to each other, and more importantly, well-connected nodes tend to be in the center of the graph. We used strength centrality (summing up the weights of edges between nodes) as a measure for importance of the node in the network. An illustration of the network can be found in the Appendix (see Figure A2 in the Appendix). We found that considering build-up elements was most central to the network, followed by detailing indicators for the category, Chain-of-Thought coding, giving a task specification, omitting context information and asking for justifications—both detailed and normal (see Table A3 in the Appendix for strength values). This suggests that they play an important role in producing high quality prompts that yield reliable results. In contrast, excluding build-up elements, using an overall decision on the category and leaving out justifications were prompting strategies that were located at the edges of the network, suggesting that they are largely irrelevant for producing good prompts.

Thus, the network analysis gave us an idea of potentially fruitful combinations of prompting strategies. As the literature, as described above, points to central prompt components (Korzynski et al., 2023) that can be supported by different prompting strategies, the question is whether our analyses reflect this in some sense. To address this in a third and last step of our analysis of the effects of combinations of prompting strategies, we also looked at the best prompt across both categories[7] in order to consider a particularly successful configuration from our tests. The reasoning behind this is that a prompt that can, in principle, be generalized could serve as a good starting point for coding, even though prompts for specific categories might achieve better reliability scores. Another advantage is that we are working with a prompt whose reliability we have already tested, rather than a hypothetical, model-based prompt, as we would if we were relying on effect sizes. The selected prompt had good reliability ($\alpha_{climate}$ = .71, $\alpha_{movement}$ = .74) and the following configuration: (1) Use chatbot as a role prompt, (2) leave out context information, (3) don't use a specification of the task, and (4) code detailed indicators for the category, (5) consider build-up elements, (6) don't use Zero-Shot Chain-of-Thought but (7) explain the steps of the analysis to the LLM through Chain-of-Thought prompting while also (8) demanding a detailed justification of the decision. Compared to this general well-performing prompt, the best category-specific prompts showed only small increases in reliability ($\alpha_{climate}$ = .76, $\alpha_{movement}$ = .78).

Considering both the network and the best prompt, we found that the presence of a prompting strategy that acts as a contextual description was important. In the case of our best prompt, role prompting seemed sufficient to achieve this goal. Regarding the instructional component of the prompt, the results can be seen as confirmation of the existence of some similarities between human coding and LLM coding. More generally, it is important to provide LLMs with sufficient information about their specific coding task, just as it is useful to get them to engage with the task in more detail and to reflect this in their answers. Specifically, both our best prompt and our network analysis showed that it was important to code detailed indicators for the category, to consider build-up elements in the codebook, to explain the steps of the analysis to the LLM through Chain-of-Thought prompting, and to require a detailed justification of the decision *before* giving the final binary answer. The choice between the more extensive Chain-of-Thought prompting and its simplified version, Zero-Shot Chain-of-Thought prompting, was clearly in favor of Chain-of-Thought prompting. In the stronger generalization, i.e., beyond the best prompt, a general task specification did not seem to be disadvantageous either, as our network analysis of the acceptable prompts (Table A3 in the Appendix) showed, even though the best prompt did not use it.

[7] We found seven additional prompts in our dataset of 41 prompts with acceptable reliability that had identical configurations across both variables (see Table A4 in the Appendix). The translated category-specific version of the best prompt can be found in the Appendix in Table A5. Table A6 shows the results of the other key evaluation metrics for this prompt and the two variables.

**Table 3.** Prediction of Krippendorff's Alpha

| | Model 1 | | | Model 2 | | | Model 3 | | | Model 4 | | |
|---|---|---|---|---|---|---|---|---|---|---|---|---|
| *Predictors* | *b* | *SE* | *p* | *b* | *SE* | *p* | *b* | *SE* | *p* | *b* | *SE* | *p* |
| (Intercept) | 0.26 | 0.00 | **<0.001** | 0.25 | 0.00 | **<0.001** | 0.34 | 0.01 | **<0.001** | 0.28 | 0.01 | **<0.001** |
| *Ground truth (ref.: $N_{coders}$)* | | | | | | | | | | | | |
| $N_{experts}$ | 0.10 | 0.00 | **<0.001** | 0.10 | 0.00 | **<0.001** | 0.10 | 0.00 | **<0.001** | 0.10 | 0.00 | **<0.001** |
| *Repetitions (ref.: 1 times)* | | | | | | | | | | | | |
| 5 times | | | | 0.01 | 0.00 | **<0.001** | 0.01 | 0.00 | **<0.001** | 0.01 | 0.00 | **<0.001** |
| *Variable (ref.: $v_{climate}$)* | | | | | | | | | | | | |
| $v_{movement}$ | | | | | | | -0.21 | 0.01 | **<0.001** | -0.13 | 0.01 | **<0.001** |
| *Role prompting (ref.: (0) none)* | | | | | | | | | | | | |
| (1) scientist | | | | | | | | | | 0.01 | 0.01 | 0.355 |
| (2) chatbot | | | | | | | | | | -0.01 | 0.01 | 0.266 |
| *Context information (ref.: (0) none)* | | | | | | | | | | | | |

| | | | |
|---|---|---|---|
| (1) general description | 0.00 | 0.01 | 0.607 |
| *Task specification (ref.: (0) none)* | | | |
| (1) general description | 0.03 | 0.01 | <**0.001** |
| *Coding strategy (ref.: (1) overall)* | | | |
| (2) detailed | 0.11 | 0.01 | <**0.001** |
| (3) detailed with limitations | -0.19 | 0.01 | <**0.001** |
| *Coding elements (ref.: (0) none)* | | | |
| (1) build-up elements | -0.01 | 0.01 | 0.091 |
| *Zero-Shot (ref.: (0) none)* | | | |
| (1) proceed | -0.01 | 0.01 | 0.265 |
| (2) think | -0.00 | 0.01 | 0.676 |
| *Chain-of-Thought (ref.: (0) none)* | | | |

| | | | | | | | | | | | | |
|---|---|---|---|---|---|---|---|---|---|---|---|---|
| (1) explanation | | | | | | | | | | 0.04 | 0.01 | **<0.001** |
| *Justification (ref.: (0) none)* | | | | | | | | | | | | |
| (1) normal | | | | | | | | | | -0.04 | 0.01 | **<0.001** |
| (2) detailed | | | | | | | | | | -0.03 | 0.01 | **<0.001** |
| **Random Effects** | | | | | | | | | | | | |
| $\tau_{00}$ | | 0.03 | | | 0.03 | | | 0.02 | | | 0.01 | |
| $\sigma^2$ | | 0.00 | | | 0.00 | | | 0.00 | | | 0.00 | |
| $N_{permutations}$ | | 1,512 | | | 1,512 | | | 1,512 | | | 1,512 | |
| $N_{observations}$ | | 6,048 | | | 6,048 | | | 6,048 | | | 6,048 | |
| Marginal $R^2$ / Conditional $R^2$ | | 0.064 / 0.885 | | | 0.065 / 0.886 | | | 0.356 / 0.886 | | | 0.576 / 0.887 | |
| AIC | | -10,515 | | | -10,580 | | | -11,212 | | | -11,950 | |

*Note.* $\tau_{00}$ = between-group random intercept variance; $\sigma^2$ = residual variance.

**Discussion**

We find that requests sent to two LLMs varied greatly across evaluation metrics (RQ1). This is consistent with previous research (e.g., Reiss, 2023) and stresses the necessity to send multiple requests and average their results to increase stability and robustness. We identified 5, 15, and 25 repetitions as good thresholds, with five repetitions being a good compromise between quality and computational cost. Other authors have made similar recommendations (e.g., Wang et al., 2023). Moreover, from a technical point of view, five repetitions have some advantages over fewer repetitions. An odd number always results in a majority, provided the number of possible categories is smaller than the number of repetitions and there are no missing values. However, since missing values do occasionally occur, five possible answers offer a greater chance that the LLM can rely on several answers in the best case, but at least one answer in the worst case.[8] Furthermore, we observed that better prompts tended to be more stable, which is important for scalability. With respect to RQ2, we found that all four factors investigated (quality of the manually coded data, repeated coding with majority decision, type of variable, prompting strategies) were important in explaining differences in the reliability of the LLM coding. To explore the interplay between prompting strategies using expert coding as ground truth and self-consistency coding, we compared prompts with acceptable reliability (Krippendorff's Alpha ≥ .67) with the remaining prompts. We found that the groups differed, with a significantly higher proportion of acceptable prompts using a detailed coding strategy, build-up elements, Chain-of-Thought prompting, and normal or detailed justifications. Looking at the most prominent combinations through network analysis and considering the best prompt across our variables as a particularly successful configuration from our tests further allowed us to examine how different combinations of prompting strategies work together, taking into consideration the theoretically derived role of central prompt components. Our results highlight that, while some individual prompting strategies may appear to reduce the reliability of LLM coding when looking at them separately, they can still lead to better outcomes when correctly combined with other strategies. Concerning the prompt components, our best prompt combines elements outlined by Korzynski et al. (2023), like the presence of contextual descriptions (role prompting) and instructional components (detailed coding of indicators, build-up elements, Chain-of-Thought, detailed justification). Based on the results of the prestudy, we see that consistent and reliable coding is possible with LLMs. Building on these findings, we deduced a pipeline that provides a systematic approach to LLM coding.

## Introducing HALC: The Hohenheim Automated LLM Coding pipeline

In our view, LLM coding should be a systematic approach that finds reliable prompts for each case, rather than randomly finding the best prompt through intensive trial and error.

[8] In fact, our tests showed that the number of final misses is highest when there is no self-consistency prompting, i.e., when only one answer is used (*M* = 0.89, *SD* = 1.71, *N* = 3,024), lower when there are three repetitions (*M* = 0.24, *SD* = 0.76, *N* = 3,024), and lowest when there are five (*M* = 0.15, *SD* = 0.56, *N* = 3,024). The differences were significant.

To this end, an LLM coding pipeline benefits from being grounded in the tradition of content analysis, validated by using a small dataset of manually coded data, while also being transparent and adaptable in the process of achieving reliable codings. Our proposed pipeline aims to fulfil these criteria. It starts with established manual coding paradigms and introduces LLM coding as an extension. In a more committed form than previous authors, we consider manual coding as ground truth against which we measure the quality of LLM coding. Our pipeline consists of the following steps:

(1) A codebook is developed conventionally based on research design, findings from literature, research questions, and methodology.

(2) Manual coding is carried out with a small (random) sample of the content that should be categorized. The results are checked for reliability using standard procedures.

(3) As soon as the reliability of the manual coding is ensured, the coding instructions in the conventional codebook are used as the basis for prompts to an LLM, which then carries out further coding.

- (3.1) An LLM is chosen and set up for coding.
- (3.2) A set of candidate prompts is chosen for evaluation (either many different prompts, a smaller subset of prompts, or a single base prompt).
- (3.3) A rule-based translation of the codebook is carried out for the LLM.

(4) The results of the manual coding are used to validate the LLM coding.

- (4.1) A reliability test is carried out using the prompts constructed in step (3).
- (4.2) If the desired reliability is not achieved, revert to step (3).

(5) As soon as the LLM codes reliably, all material can be fully coded by the LLM.

For steps **(1)** and **(2),** an existing codebook can be selected. Alternatively, new codebooks can be designed based on established methods (e.g., Früh, 2017; Neuendorf, 2002; Rössler, 2017). Since we consider human coders to be ground truth, there is no need for LLM consideration in these steps. The manual codebook must be evaluated to ensure its quality and that human coders can agree on how to code and produce reliable results. Once the codebook is ready to be used for coding, researchers have some choices to make in step **(3)**. First, a suitable LLM must be chosen **(3.1)**. We recommend using a recent and potent LLM that ensures sufficient linguistic capabilities and general knowledge. We highly recommend using self-hosted open-source models whenever possible, as they bypass potential privacy issues and ensure reproducibility by giving total control to the scientist. Second, suitable prompting strategies must be chosen **(3.2)**. We recommend starting with a single potent base prompt such as the one we found in the prestudy. Alternatively, a set of prompting strategies can be combined into a set of promising prompts. Ultimately, all possible known prompting strategies could be combined, depending on available resources and requirements. We recommend increasing the number of prompt candidates systematically and iteratively until a prompt with the desired reliability is found. Generally, prompts should adhere to the four essential components of effective prompts, as described above: context description, instruction, data to be processed, and output format. Third, a rule-based translation of the codebook is carried out **(3.3)**. To this end, our recommendations from the review of literature on prompt engineering and manual content analysis can be considered (see chapter *Guidelines for the rule-based translation of the codebook for the systematic construction of prompts*). In some cases, this needs to be aligned with the prompting strategies. Once step **(3)** is finished, the prompts are sent to an LLM and the results are evaluated **(4)**. Based on our empirical findings, we recommend making (at least) five repetitions and using majority decisions to increase consistency and reliability. In step **(4.1),** the LLM responses are validated regarding reliability using the manually coded data. Steps **(3)** and **(4)** are intertwined, since systematic iterations may be necessary if the tested prompts are not sufficiently reliable **(4.2)**: In this case, the process can revert to refining the codebook translation **(3.3)**, incorporating other prompting strategies **(3.2)**, or switching to a different LLM model **(3.1)**. These iterations must be kept at a minimum to avoid overfitting. To ensure transparency, all changes should be documented based on our guidelines for a rule-based translation of the codebook. For example, changes can be recorded according to their intent to improve the prompt's completeness or explicitness. If more extensive iterations are necessary, additional or new ground truth data needs to be coded for LLM validation. Once the coding is deemed reliable, it can be applied to the entire dataset, excluding test data **(5)**.

To evaluate our proposed pipeline, we applied HALC to new datasets to test its suitability for different topics, variables, and languages. As part of this process, we also tested how it behaves when different models are applied.

## Main study: Application of the HALC pipeline across thematical and linguistic contexts

### Method

*Data and reliability*

We evaluated the applicability of our pipeline on two datasets. These were chosen because they form the basis of published studies, are partly publicly available, have reported relatively high levels of reliability, and cover different topics and languages. This enabled us to test the effectiveness of HALC in achieving reliable coding across different conditions.

The first dataset is the publicly available Parties' Immigration and Integration Positions (PImPo) dataset, which analyses migration discourse within national party manifestos across diverse political systems (Lehmann & Zobel, 2018a, 2018b). It consists of 235,132 quasi-sentences from 14 countries. Our analysis uses a subset of 115,182 quasi-sentences, encompassing the complete available longitudinal records for the USA, Spain, Germany, Austria, and Switzerland (1998 – 2014). We chose two variables coded in the original study: the *selection question*, a binary indicator of whether a quasi-sentence addresses migration at all, and the binary *topic question* that distinguishes between immigration, understood as the regulation of entry, admission, and border control, and immigrant integration, understood as the social, economic, cultural, and political inclusion of immigrants already residing in the country. Of the 115,182 quasi-sentences in our sample, 3.9% were coded as addressing immigration or immigrant integration at all, indicating that migration-related content is relatively sparse in the corpus. Among these migration-related quasi-sentences, 45.4% were classified as referring to immigration and 54.6% as referring to immigrant integration. Lehmann and Zobel (2018b) report high median Krippendorff's Alpha values between .86 and 1.00 for the different countries regarding the selection question (see Table 4) and values between .68 and .79 regarding the topic question (see Table 5). The PImPo dataset therefore provides a good basis for assessing the applicability of HALC under high-sparsity conditions, across different languages as well as different varieties of the same language.

While the first dataset stems from a political context, the second relates to business communication. We used a corpus of German-language annual reports of the 30 companies listed in the DAX stock index in 2014 (DAX-30; Thoms et al., 2020).[9] The corpus consists of 2,902 text paragraphs that were manually coded to discern communicative tone while also considering relatedness to the company. Thoms et al. (2020) coded five categories: no company reference, neutral, positive, negative, and ambivalent company references. In total, 2.8% of the paragraphs did not reference the company, 26.8% were neutral, 39.5% were positive, 10.6% were negative, and 20.4% were ambivalent. The authors report a Krippendorff's Alpha of .79, indicating a high reliability despite the complexity of the coding task, which involves a high degree of semantic density and specialized vocabulary and thus requires high levels of coding discipline for accurate classification. The DAX-30 dataset provides a good basis for assessing the applicability of HALC in a technical communication context with a complex multi-label classification that includes an assessment of tone.

*Technical infrastructure*

To evaluate architectural variance, we deployed two state-of-the-art open-weights LLMs of comparable size via a local inference engine (vLLM[10]). We used Qwen 2.5 32B[11], an instruct model, and Gemma 4 31B[12], a reasoning model. To ensure methodological accessibility, both models used 4-bit quantization, allowing execution on local hardware without compromising semantic grounding. We applied the self-consistency method as explained in the prestudy, executing five independent requests per unit of analysis at a temperature of 0.7 to permit necessary interpretive variance. Final classifications were derived through majority decision (at least 3 of 5 codings). In the case of the DAX-30 dataset, due to the category scheme used, any unit that did not reach an absolute majority was classified as ambivalent/unclear. We executed a total of approximately 1.2 million automated coding requests across five repetitions, two LLMs and tasks for the PImPo dataset and 29,020 automated requests for the DAX-30 dataset.

[9] This corpus is not publicly available due to copyright restrictions, but it can be provided by the authors for academic purposes upon request.
[10] https://docs.vllm.ai/en/latest/
[11] https://huggingface.co/Qwen/Qwen2.5-32B-Instruct-GPTQ-Int4
[12] https://huggingface.co/cyankiwi/gemma-4-31B-it-AWQ-4bit

*Construction and validation of prompts*

A central element of the application of HALC is the construction of adequate prompts for the coding tasks at hand. This includes the selection of prompting strategies (Step 3.2), the rule-based translation of the codebook (Step 3.3), and the validation of the prompts (Step 4), including possible prompt refinement iterations. Regarding Step 3.2 and based on our prestudy, we used a prompt that combined role prompting, coded detailed indicators for the category, considered build-up elements (where applicable), used Chain-of-Thought prompting and demanded a detailed justification of the decision. We proceeded with the rule-based translation of the human codebook for the LLM coding according to our proposed guidelines (Step 3.3) and validated the prompts (Step 4). This included two rounds of refinement for the PImPo tasks and one round for the DAX-30 task. These refinements focused on improving either completeness (e.g., adding missing indicators) or explicitness (e.g., clarifying semantic boundaries). The prompt for coding the selection question in the PImPo dataset was validated using up to 200 migration-related sentences per country, as well as twice as many unrelated sentences (n = 1,686). Reliability calculations used weighting to match the original high-sparsity distribution. The prompt for coding the topic question was developed using a sample of 562 sentences. For the DAX-30 dataset, all paragraphs were used directly.

For the PImPo dataset, we started with a translation of the original categories (Lehmann & Zobel, 2018b, 2018c) into two prompts, one for the selection question and one for the topic question. In both cases, warnings originally provided to human coders were mapped as strict negative constraints within the prompt instructions. Furthermore, all texts were coded using the same English language prompts. This methodological choice eliminated the confounder of prompt translation in multilingual scenarios, taking advantage of the multilingual capabilities of LLMs. The first refinement round for the selection question prompt addressed corpus sparsity by incorporating explicit codebook-derived exclusion criteria (e.g., general civics, indigenous groups, or emigration) and adding narrative indicators to structure the prompt. The second round sharpened the theoretical boundary between “entry” and “membership”; we implemented the codebook-derived “conduct vs. status” rule for deportation and realigned “naturalization” as an integration process to match the codebook. After this second round, we achieved a Krippendorff’s Alpha of 0.73 using the Qwen model, which was considered acceptable for coding the rest of the material. Compared to the selection question, the translation of the topic question primarily aimed at distinguishing between immigration and integration. This was achieved by adding more information and decision rules from the codebook. After the second refinement round, we achieved a Krippendorff’s Alpha of 0.69 using Qwen, which was considered acceptable for coding the rest of the material.

The DAX-30 dataset required mapping a hierarchical decision tree as well as five distinct labels, four of which represent communicative tones. We translated the original category into a sequential reasoning path (reference to company > neutrality > evaluation), and integrated domain-specific word lists closely following the study’s codebook. To ensure the model respected financial semantics, we included explicit direction rules (e.g., classifying an increase in liabilities as negative sentiment). In a single refinement round, we integrated 192 domain-specific keywords also used in the original codebook and introduced a directional rule to ensure the model correctly classified the inverse relationship between increasing liabilities (e.g., costs or debts) and corporate success. Moreover, we added a final re-evaluation step to the end of the prompt to enforce the hierarchical priority of the categories (reference to company > neutral > positive or negative > ambivalent). Since we used all the paragraphs directly, the Krippendorff’s Alpha value after this round is the same as the final reported value.

The full documentation of the final prompts for the three variables is provided in Table A7 in the Appendix.

## Results

### *PImPo dataset*

Regarding the selection question, our results suggest that both models performed well overall in identifying migration-related content within a corpus dominated by general political discourse, as indicated by Krippendorff’s Alpha values of .74 and .75 (see Table 4). In the context of quantitative content analysis, these reliability scores indicate that using our pipeline to construct prompts based on a codebook developed for manual content analysis results in LLM codings that are reasonably reliable given the original human crowd-coding effort. Our results also demonstrate that the English-language prompt can be used for reliable multilingual codings with some exceptions. While the models maintained substantial agreement with the human codings across Spanish and German manifestos, the results regarding the USA (EN) subset highlighted a functional boundary. A qualitative review of these discrepancies revealed the following issue: the model frequently classified general discussions of American national values, diversity, or border security as migration-specific, whereas human coders in the original PImPo dataset applied a stricter, more exclusive definition. This finding emphasizes that researchers must remain aware of how cultural narratives can influence automated coding. Finally, our results also show interpretive sensitivity when navigating unique national registers. The comparison of Austria (AUT), Germany (GER), and Switzerland (SUI) allowed us to study the suitability of our prompt to code texts using different varieties of the same language. In the Switzerland (SUI) subset, which features highly specific political terminology related to direct-democratic initiatives, both models achieved their peak performance ($\alpha = .83$). This high reliability suggests that the logic mapped in Step 3.3—which emphasized indicators like border management and residency rights—is robust enough to discern context-dependent meaning without requiring special keyword dictionaries.

**Table 4.** Reliability of the selection question coding across languages and models

| | | | Qwen (Instruct) | | | | | Gemma (Reasoning) | | | | |
|---|---|---|---|---|---|---|---|---|---|---|---|---|
| **Country** | **N** | **Orig. α** | **α** | **Acc** | **Pr** | **Re** | **F1** | **α** | **Acc** | **Pr** | **Re** | **F1** |
| AUT (DE) | 19,909 | 0.92 | .77 | .98 | .81 | .75 | .78 | .76 | .98 | .72 | .84 | .78 |
| ESP (ES) | 40,074 | 0.91 | .70 | .99 | .70 | .72 | .71 | .67 | .98 | .60 | .78 | .68 |
| GER (DE) | 41,149 | 0.86 | .74 | .98 | .89 | .65 | .75 | .78 | .98 | .82 | .76 | .79 |
| SUI (DE) | 5,198 | 0.95 | .83 | .98 | .89 | .79 | .84 | .83 | .98 | .84 | .86 | .85 |
| USA (EN) | 8,852 | 1.00 | .65 | .98 | .55 | .81 | .66 | .62 | .98 | .48 | .93 | .63 |
| **OVERALL** | **115,182** | **—** | **.74** | **.98** | **.80** | **.71** | **.75** | **.75** | **.98** | **.72** | **.80** | **.76** |

*Note.* N = unique sentences; Orig. α = estimated Krippendorff’s Alpha from Lehmann & Zobel (2018b); α = Krippendorff’s Alpha (nominal); Acc = Accuracy; Pr = Precision; Re = Recall; F1 = F1-score.

The second variable coded for the PImPo dataset was the topic question. This task increased the interpretive difficulty compared to the selection question, as it required the model to identify priorities within a shared thematic context. As shown in Table 5, overall reliability scores remained stable, with the reasoning model Gemma achieving a Krippendorff’s Alpha of .71, compared to the instruct model Qwen with a Krippendorff’s Alpha of .67. While both architectures demonstrated a comparable ability to recognize the thematization of migration (selection question), a small but consistent discrepancy in topic discernment emerged between immigration and immigrant integration (topic question). Here, Gemma’s reasoning capability yielded a more stable alignment with human experts across most national subsets. Again, the English-language prompt led to comparable results across different languages. However, the Spanish (ESP) subset fell behind already liberal reliability thresholds of Krippendorff’s Alpha with values of .53 (Qwen) and .61 (Gemma). A review of these discrepancies suggested that the Spanish political landscape—characterized by discussions where border management and residency rights are frequently conflated within the context of European solidarity—presented a greater linguistic diversity than was reflected in the original codebook, which was based on a more basic „entry vs membership“ distinction. Within the German speaking countries (AUT, GER, SUI), the prompt constructed using HALC demonstrated a high degree of alignment with the human codings. The reasoning model reached Krippendorff’s Alpha values between .73 and .75, exceeding the median original human crowd-coding benchmarks (Orig α =.68 to .74). As with the coding of the selection question, this finding suggests that the definitions incorporated in the prompt worked well to code texts from three distinct political systems that share the same linguistic root. In this specific coding task at least, automated coding respected the functional context of the manifestos rather than relying solely on linguistic cues.

**Table 5.** Reliability of the topic question coding across languages and models

| | | | Qwen (Instruct) | | | | | Gemma (Reasoning) | | | | |
|---|---|---|---|---|---|---|---|---|---|---|---|---|
| **Country** | **N** | **Orig. α** | **α** | **Acc** | **Pr** | **Re** | **F1** | **α** | **Acc** | **Pr** | **Re** | **F1** |
| AUT (DE) | 998 | 0.74 | .73 | .86 | .87 | .87 | .86 | .75 | .87 | .88 | .88 | .87 |
| ESP (ES) | 965 | 0.79 | .53 | .77 | .78 | .77 | .77 | .61 | .81 | .81 | .81 | .80 |
| GER (DE) | 1,951 | 0.68 | .69 | .87 | .88 | .83 | .85 | .73 | .88 | .89 | .85 | .87 |
| SUI (DE) | 387 | 0.71 | .70 | .85 | .87 | .85 | .85 | .74 | .87 | .88 | .87 | .87 |
| USA (EN) | 162 | 0.72 | .68 | .84 | .84 | .84 | .84 | .68 | .84 | .84 | .84 | .84 |
| **OVERALL** | **4,463** | **—** | **.67** | **.84** | **.86** | **.83** | **.84** | **.71** | **.86** | **.87** | **.85** | **.86** |

*Note.* N= sentences identified as migration-related by human experts; Orig. α = estimated Krippendorff's Alpha from Lehmann & Zobel (2018b); α = Krippendorff's Alpha (nominal); Acc = Accuracy; Pr = Precision; Re = Recall; F1 = F1-score.

*DAX-30 dataset*

The second dataset used in our main study coded five categories in a business communication context and included an evaluation of the communicative tone. The task was set up as a multi-label classification case where the models had to analyze business vocabulary and apply hierarchical decision rules (e.g., distinguishing factual company status from evaluative interpretation), as outlined above concerning the rule-based translation of the codebook in Step 3.3 of HALC. As shown in Table 6, the chance-corrected reliability reached only a Krippendorff's Alpha of .51 using Qwen and of .69 using Gemma.

**Table 6.** Reliability of the tone coding across models

| **Model Architecture** | **α** | **Acc** | **Pr** | **Rec** | **F1** |
|---|---|---|---|---|---|
| Qwen (Instruct) | .51 | .65 | .67 | .67 | .63 |
| Gemma (Reasoning) | .69 | .77 | .75 | .79 | .76 |

*Note.* α = Krippendorff's Alpha (nominal); Acc = Accuracy; Pr = Precision; Re = Recall; F1 = F1-score.

Going into more detail, the reasoning model was particularly effective in stabilizing the most interpretive categories (see Table 7). This suggests that higher reasoning depth is required for the model to effectively apply the exclusion rules formulated for the prompt in Step 3.3. The most pronounced differences occurred in the discernment of neutral content. Furthermore, the reasoning model handled the high semantic density of the ambivalent category better than the instruct model. This observation indicates a more consistent application of the hierarchical decision tree when parsing paragraphs containing conflicting facts. Overall, using the same prompt, the reasoning model outperformed the instruction model. This demonstrates that factors other than prompt formulation influence coding reliability, as outlined in our literature review. However, it also hints at the possibility of switching to a different model if other measures do not significantly improve reliability, which is also a consideration in our pipeline (Step 4.2 and 3.1).

**Table 7.** Category-level Krippendorff's Alpha (α) comparison of the tone coding across models

| Category | N | α Gemma (Reasoning) | α Qwen (Instruct) |
|---|---|---|---|
| No reference to company | 80 | .80 | .61 |
| Neutral | 778 | .71 | .44 |
| Positive | 1,146 | .72 | .56 |
| Negative | 307 | .66 | .58 |
| Ambivalent | 591 | .60 | .48 |

## Discussion

As a first application of the HALC pipeline, the aim of our main study was to answer RQ3:

> *Does applying the pipeline lead to reliable coding across different datasets, variables, languages, and LLMs?*

Drawing on the prompting strategies identified in the prestudy and existing codebooks, we used HALC to construct prompts for three coding tasks: two binary classification tasks related to topic identification and one multi-label classification, which includes an assessment of tone. These tasks cover two different thematic contexts (political and business communication), different document types (election manifestos and annual reports), different languages (English, German, and Spanish for the manifesto dataset, with an additional comparison of different German-language contexts), and different coding units (sentences and paragraphs). The chosen models, Qwen and Gemma, allow for a first comparison of the behavior of an instruct and a reasoning model when using the same prompts. Although HALC offers several customization options during the construction process, we chose to keep the prompting strategies consistent, unlike in the prestudy. Instead, we used the rule-based translation of the codebook as a means of adjustment. Thus, for this test of HALC's applicability, we did not aim for the best possible result, for which further tuning options could have been used. Instead, we kept certain parameters constant and focused primarily on content-based adjustments. This made it possible to compare the same basic prompt structure across the different tasks.

Our main study offers several insights. Although this is more of an anecdotal account, using HALC leads to a significantly more rigorous approach to prompt construction. This is primarily reflected in fewer, more systematic adjustment loops that can clearly be named according to their aim (e.g., increasing completeness or explicitness). As mentioned, we deliberately did not address all possible adjustment parameters and focused on the rule-based translation of the codebook, which streamlined the process. However, our results show that using a basic prompt structure adapted for different variables with HALC overall yields acceptable results for various coding tasks and models. We would like to emphasize that our ultimate goal is not to create a prompt that is a one-size-fits-all solution. The use of HALC is always about developing prompts tailored to the specific task at hand while following a systematic approach. However, we interpret our results as indicating that acceptable outcomes can be achieved with relatively little effort if certain fundamental principles are taken into account. This is also reflected in the fact that LLMs seem to be able to navigate multilingual corpora using a single set of English-language instructions, which simplifies the automated coding process and minimizes translation bias. In our view, our pipeline helps avoid making prompt construction an exercise in linguistic trial and error. It also contributes to the transparency of the process by reducing the researcher degrees of freedom. Furthermore, when we compared the reliability of the full dataset to that during prompt construction, we observed only minor discrepancies. This indicates that our approach of first constructing reliable prompts based on a sample and then applying them to the rest of the material (Step 5) is effective.

Nevertheless, differences in performance across different coding tasks became evident. This underscores that many different factors ultimately influence the reliability of LLM coding. For example, we observed differences in coding across different cultures and languages. Regarding our decision to code all texts with one English-language prompt, one way to increase reliability would be to translate and adapt the prompts for a better fit with the language and political context. Similarly, the complexity of the coding task was a decisive factor. Here, we also identified initial indications of differences in the performance of various model architectures. For complex variables, such as our hierarchical coding of tone, reasoning models appeared to have an advantage compared to instruct architectures, but this requires further investigation. For now, we conclude that a final adjustment could consist in changing the LLM (Step 4.2 and 3.1) if the other steps fail to achieve acceptable reliability.

## General discussion

In this study, we aimed to demonstrate how LLMs can be utilized in a systematic and transparent way to produce reliable codings in content analyses. Our study contributes to the scientific landscape in several ways. First, we synthesized insights from the literature, identified challenges in the use of LLMs, discussed influencing factors, outlined existing frameworks for collaboration between human coders and LLMs, discussed existing prompting strategies, and concluded with general guidelines for a rule-based translation of codebooks into prompts. Second, in a prestudy, we tested the consistency of repeated LLM codings and examined factors that influence the quality of the codings. We found that repeated prompting stabilizes coding and increases quality. We also found that the quality of the manually coded data, repeated coding with majority decision, the type of variable, and prompting strategies influence coding quality. We presented a combination of prompting strategies that yielded good results in the prestudy. Third, we combined the findings from the literature and our prestudy into the HALC pipeline that provides a systematic approach to LLM coding. Our incentive was to address technical challenges affecting the consistency and reliability of LLM coding, and epistemic challenges concerning the validity and half-life of research findings in a pipeline for the systematic combination of LLMs and human coders based on established scientific standards for content analyses and recent findings from the field of computer science on prompt engineering. Finally, we applied the HALC pipeline to new datasets covering different topics, variables, and languages. In doing so, we demonstrated that using HALC, prompts can be deduced in a systematic and transparent way, resulting in overall acceptable reliability scores.

In addition to these key findings, we provide further practical advice for researchers on using LLMs in content analyses. When choosing an LLM, we recommend the use of local language models over API-based solutions (e.g., *ChatGPT*) for several reasons. First, local models offer superior reproducibility. API-based models are frequently updated or deprecated, limiting long-term comparability. In contrast, once a local model is validated for a task, it can be reused indefinitely. Second, local use minimizes data privacy concerns. Since no data leaves the research environment and no training occurs, sensitive information can be processed. And third, while commercial APIs may offer higher efficiency due to extensive optimization, we find that local models like *Mistral NeMo* or *Gemma4* can be sufficiently reliable for automated coding tasks. Generally, the open-source nature of LLMs available for local use provides greater transparency and is therefore more suitable for research projects. While it is true that running an LLM locally is challenging for non-experts, there are tools that can facilitate this, like *ollama* (2023). Even if no GPU and only basic hardware is available, slow LLM requests are usually still possible. Using this solution is elegant because the models can then be used via a local API in statistical software like R or with Python. However, we still acknowledge the utility of cloud-based LLMs as these often provide state-of-the-art performance and ease of use. If researchers prefer these, HALC is also compatible with such options.

We also encountered some technical challenges that should be considered when using LLMs for automated coding. Models may fail to follow instructions accurately, resulting in problems such as generating incorrect or malformed JSON, or inconsistently applying coding schemes. This often requires repeated requests or additional validation steps and can add significant computational overhead. In our system, we automatically re-request failed coding attempts until the LLM provides a valid response or reaches a repetition threshold.

As mentioned earlier, the field of LLMs is undergoing constant change, making it difficult to keep up. In the course of our research, studies have been published that present new prompting strategies. One example is the recently published Chain-of-Draft strategy (Xu et al., 2025), which improves Chain-of-Thought. Strategies like this, or other known strategies like Few-Shot-Learning (Brown et al., 2020), can be easily incorporated into our presented pipeline and tested for effectiveness and interaction with other prompting strategies potentially improving results. In general, by investigating newer and better strategies, or by aligning and combining similar strategies, fewer strategies need to be considered. In other words, the grid-search-like structure of our approach in the prestudy, which results in a combinatorial increase in complexity following $O(n^d)$ growth, can be mitigated by, for example, testing only one prompting strategy for each prompt component. As we observed in our tests, redundancies such as testing combinations of two different types of contextual descriptions (in our case, role prompting and context information) or combining the more extensive Chain-of-Thought prompting with its simplified version, Zero-Shot Chain-of-Thought, were not necessary. Choosing one of several similar strategies should be sufficient, as we also found in our main study. In the future, our pipeline could also be extended to work as a multi-LLM approach, where prompting strategies are optimized for different LLMs. Coding performance could potentially be improved by leveraging multiple different LLMs as coders, as has been previously proposed by other researchers (Xiong et al., 2023). Here, we can refer to the results of our main study regarding the coding of the selection question. These showed that the recall and precision of different models sometimes exhibited contradictory trends (see Table 4). This suggests that models could potentially complement one another.

## Limitations

One limitation is that it is beyond our scope to provide a detailed, systematic review of the current literature regarding LLM coding. We did our best to reflect current discussions and developments, but the field is evolving at a very rapid pace. Furthermore, many findings—particularly in the area of prompt engineering—are published in gray literature, making them even more difficult to systematize. Regarding the prestudy, we tested only two LLMs. At this step, none of our models were reasoning models. Nonetheless, as we demonstrate in our main study, the HALC pipeline can be applied to any category or LLM. Furthermore, although we investigated various categories in our prestudy and main study, it is still unclear how LLM coding will behave when confronted with multi-point scales. Coding ordinal, continuous, or complex multi-label scales, or other category types, as well as hierarchical coding schemes, is theoretically possible but would introduce additional challenges and issues. This is further complicated by asking LLMs for multiple codings at once. We therefore recommend that future studies examine additional types of scales, levels of conceptual complexity, and coding tasks to further test LLM coding capabilities. There is also room for improvement regarding reliability of the coding. Although our prestudy and main study achieved values above the acceptable threshold (.67) in most cases, these may not be sufficient in some research situations. However, the HALC pipeline offers options for iteration to theoretically achieve better reliability values. Moreover, investigating interactions between prompting strategies in depth presents a further avenue for future research by focusing on a more systematic approach, since our exploratory analyses can only provide first insights. This could include studying interactions with dedicated experimental designs. Finally, even though our method is designed to simplify automated coding, it still requires a significant amount of computational power. This is particularly due to the multiple rounds of coding—which are essential for stable and reliable results—and the specific LLM selected. We are already seeing that larger models can produce better coding results, especially in situations where categories are difficult and abstract. However, this must always be balanced against the available resources.

## Conclusion

Our work shows the great potential of using LLMs to reliably, verifiably and reproducibly code data in the social sciences and in general. LLMs can be adapted to different tasks; the use of natural language prompts potentially reduces barriers for non-technical researchers. However, working with LLMs can be challenging. We present HALC—a versatile pipeline that can be applied to combine human coders and LLMs. A key feature of our proposed pipeline is to systematically find reliable prompts that are good enough, rather than randomly finding the best prompt through intensive trial and error. To this end, it is grounded in the tradition of content analysis, validated by using manually coded data, while also being transparent and adaptable in the process of achieving reliable codings. We highly encourage other researchers to build on top of our pipeline and to investigate its generalizability, reliability, and performance with their own LLMs, datasets, and scales. We believe that the proposed pipeline will make research easier without compromise.

**Appendix**

**A1.** Rule-based translation of the codebook for the prompts and prompt construction

Using the variable $v_{climate}$ (thematizing climate change) as an example, we illustrate the process of prompt construction.

The original codebook was developed for a project analyzing German articles and corresponding user comments on the climate movements *Fridays for Future* and *Last Generation*. Since the coding at both levels was related, the original category descriptions were much more detailed at the article level than at the comment level. The coders were made aware of this peculiarity in the construction of the codebook and were instructed to adapt the coding to the text type. For example, articles allow for more detailed explanations, while comments are often shorter and require more reading between the lines. In addition, readers may express opinions in their comments that differ from the prevailing societal and scientific consensus, such as questioning the veracity of climate change. This is highly unlikely for journalists writing for the type of media analyzed in the project.

The category at the article level was the starting point for the coders. At the comment level, they were given a shortened version of the category but were instructed to apply these rules more liberally at the comment level as well. As a result, the category at the article level was also the starting point for the prompt construction.

The original German language category can be translated as follows

*ai1_climate: Thematizing climate change (article level)*

*Here we code whether the article discusses climate change and its consequences in general. This includes the discussion of climate-conscious actions. This general thematization of climate change must be distinguished from thematizing the movement's goals (category ai2_goal) (e.g., when the 1.5 degree target is called for). The difference can also be seen in the language used (see category ai2_goal). Simply mentioning the terms "climate change" or "climate protection" is not enough. Instead, such a general thematization of the issue can be recognized by, for example*

- *Explaining what climate change is;*
- *Explaining what is known about climate change (e.g., based on research);*
- *Explaining what evidence there is for climate change or what makes it noticeable (e.g., forest fires, increasing extreme weather events, rising average temperatures, droughts, floods);*
- *Explaining what measures are urgently needed or could help to prevent or mitigate climate change or its consequences (e.g., by addressing basic recommendations for action, including at the political or societal level).*

*An additional condition is that the relevant information must be supported by the (proven) expertise of the person making the statement, by references to reputable sources (e.g. studies or reports), or by references to specific events. For this study, this means*

- *If only an (unsubstantiated, not further verifiable) statement is made by members of the movement (e.g. a blanket statement such as "the whole world is burning"), code 0.*
- *If members of the movement refer to specific sources or events in their statement (e.g., by referring to reports of expert committees, or by referring to a specific wildfire event rather than a general statement), code 1.*
- *If a named expert (who is not part of the movement) makes a statement, e.g. the journalist refers to specific experts and their findings or names specific events, code 1.*

*0 does not occur*
*1 occurs*

The main part of the category is the definition of what counts as thematizing climate change, including examples or indicators that illustrate this. As can be seen, the original category already included a list of relevant components and examples of indicators to help coders identify this. While the first part was relevant for the article and the comment level, the second part, which refers to the requirement of expertise of the actors making a statement, was only relevant for the article level, as this was not expected to be regularly found in a reader comment. Applying the guidelines derived from the literature on prompt engineering (completeness, conciseness, comprehensibility, clarity, explicitness, and structure), this main part of the category was translated as follows

*1) Explanation of climate change (e.g., long-term changes in temperatures and weather patterns). 2) Causes of climate change (e.g., pollution, greenhouse gas emissions, deforestation, global warming). 3) Effects and consequences of climate change (e.g., wildfire, extreme weather events, temperature increases, droughts, floods). 4) Individual, social, or political actions to prevent or mitigate climate change and its consequences (e.g., not eating meat, not flying, increasing renewable energy, climate protection, environmental protection).*

Further references to the overall context of the coding task and the specific way of coding it, which are central elements of categories and prompts, were included in different ways depending on the prompting strategies used. The following table shows the construction of a sample prompt using several of the prompting strategies reported in our study:

| **Prompt text** | **Prompting strategy or prompt component** |
|---|---|
| You are a German chatbot. | Role prompting (2) chatbot |
| You analyze reader comments under German articles about climate protest movements. | Context information (1) general description |
| You code whether reader comments address climate change. | Task specification (1) general description |
| Decide for each reader comment whether at least one of the following characteristics is present. | Coding strategy (1) overall |
| 1) Explanation of climate change (e.g., long-term changes in temperatures and weather patterns). 2) Causes of climate change (e.g., pollution, greenhouse gas emissions, deforestation, global warming). 3) Effects and consequences of climate change (e.g., wildfire, extreme weather events, temperature increases, droughts, floods). 4) Individual, social, or political actions to prevent or mitigate climate change and its consequences (e.g., not eating meat, not flying, increasing renewable energy, climate protection, environmental protection). | Translated category description from the codebook |
| Think step by step. | Zero-Shot (2) think |
| 1) Read the category description and the reader's comment. 2) Based on the category description, think about whether the reader's comment is about climate change. 3) Type the appropriate answer. | Chain-of-Though (1) explanation |
| For the comment, indicate whether at least one of the characteristics is present (true) or not (false). | Coding strategy (1) overall |
| Justify your decision. | Justification (1) normal |
| Answer using the following pattern in json format: {'reason': string, 'climate_change': boolean}. | Output format |
| Here is the comment you should analyze: | Announcement of the comment text |

The example does not include the use of build-up elements (prompting strategy: coding elements).

**Table A1.** Correlations between mean and variance of key evaluation metrics with five repetitions of 100 prompt permutations for *Mistral 7B* and *Mistral NeMo*

| **Evaluation metric** | **Mistral 7B** | | **Mistral NeMo** | |
|---|---|---|---|---|
| | ***r*** | ***p*** | ***r*** | ***p*** |
| Accuracy | -0.39 | < .001 | -0.40 | < .001 |
| F1 Score | -0.31 | .002 | -0.46 | < .001 |
| α | -0.29 | .004 | -0.40 | < .001 |
| Precision | 0.09 | .348 | 0.05 | .592 |
| Recall | -0.11 | .298 | -0.16 | .108 |

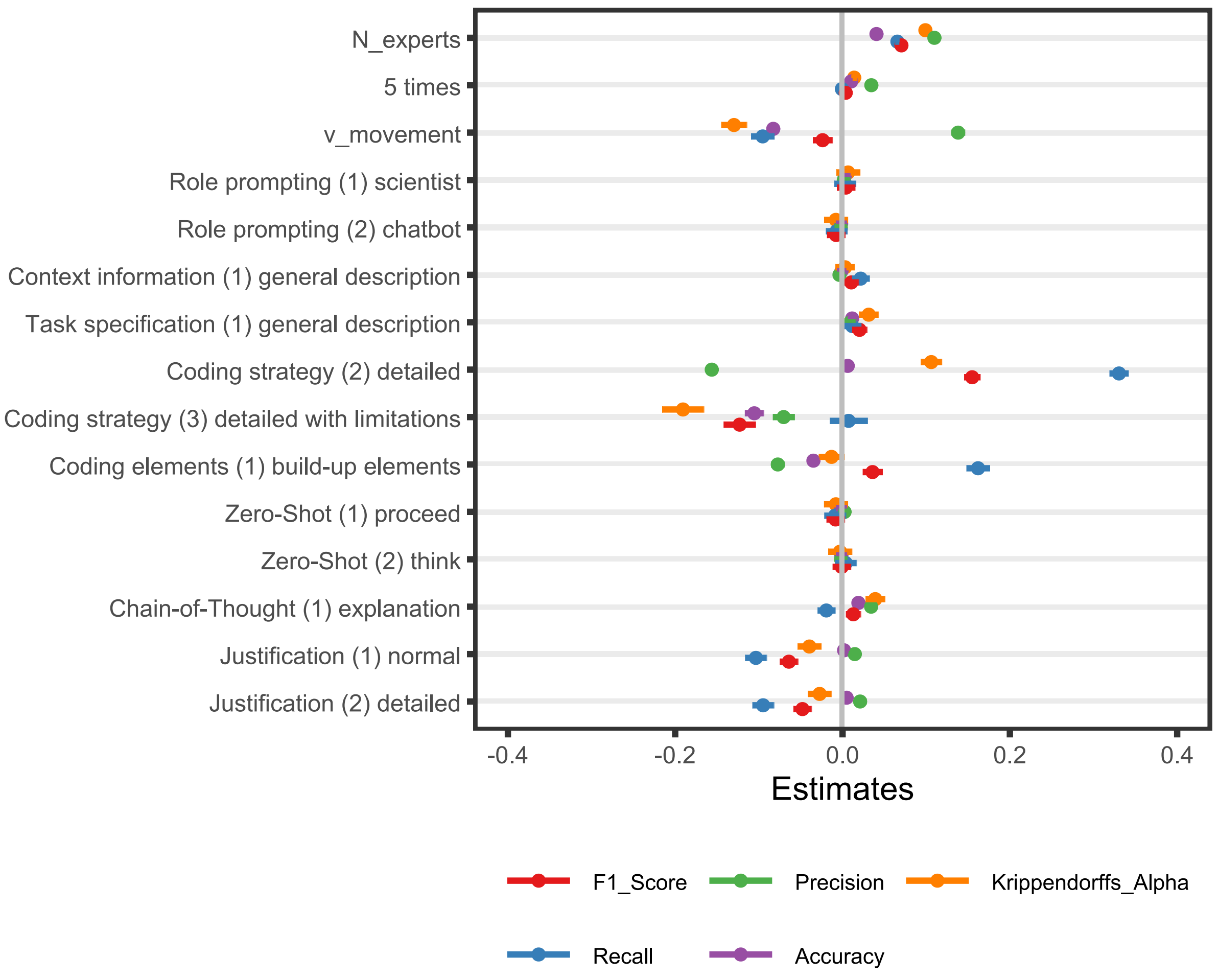

**Figure A1.** Prediction of Krippendorff's Alpha, accuracy, precision, recall, and F1 Score by different influencing factors in comparison

**Table A2.** Comparison of prompts with acceptable (≥ .67) and unacceptable Krippendorff's Alpha values

| **Prompting strategy** | **Quality** | | $\chi^2$ | *df* | *p* | *V* |
|---|---|---|---|---|---|---|
| | **acceptable** ***n (%)*** | **unacceptable** ***n (%)*** | | | | |
| *Role prompting* | | | | | | |
| (0) none | 10 (24.4%) | 494 (33.6%) | 1.55 | 2 | .460 | .03 |
| (1) scientist | 16 (39.0%) | 488 (33.2%) | | | | |
| (2) chatbot | 15 (36.6%) | 489 (33.2%) | | | | |
| *Context information* | | | | | | |
| (0) none | 24 (58.5%) | 732 (49.8%) | 0.90 | 1 | .342 | .03 |
| (1) general description | 17 (41.5%) | 739 (50.2%) | | | | |
| *Task specification* | | | | | | |
| (0) none | 16 (39.0%) | 740 (50.3%) | 1.60 | 1 | .205 | .04 |
| (1) general description | 25 (61.0%) | 731 (49.7%) | | | | |
| *Coding strategy* | | | | | | |
| (1) overall | 1 (2.4%) | 647 (44.0%) | 51.51 | 2 | < .001 | .18 |
| (2) detailed | 40 (97.6%) | 608 (41.3%) | | | | |
| (3) detailed with limitations | 0 (0%) | 216 (14.7%) | | | | |
| *Coding elements* | | | | | | |
| (0) none | 1 (2.4%) | 647 (44.0%) | 26.44 | 1 | < .001 | .14 |
| (1) build-up elements | 40 (97.6%) | 824 (56.0%) | | | | |
| *Zero-Shot* | | | | | | |
| (0) none | 16 (39.0%) | 488 (33.2%) | 0.65 | 2 | .722 | .02 |
| (1) proceed | 12 (29.3%) | 492 (33.4%) | | | | |
| (2) think | 13 (31.7%) | 491 (33.4%) | | | | |
| *Chain-of-Thought* | | | | | | |
| (0) none | 7 (17.1%) | 749 (50.9%) | 16.95 | 1 | < .001 | .11 |
| (1) explanation | 34 (82.9%) | 722 (49.1%) | | | | |
| *Justification* | | | | | | |
| (0) none | 1 (2.4%) | 503 (34.2%) | 18.25 | 2 | < .001 | .11 |
| (1) normal | 19 (46.3%) | 485 (33.0%) | | | | |
| (2) detailed | 21 (51.2%) | 483 (32.8%) | | | | |

*Note*. Values for every prompting strategy represent column percentages within acceptable ($n = 41$) and unacceptable ($n = 1,471$) prompts. Effect size is measured using Cramér's $V$.

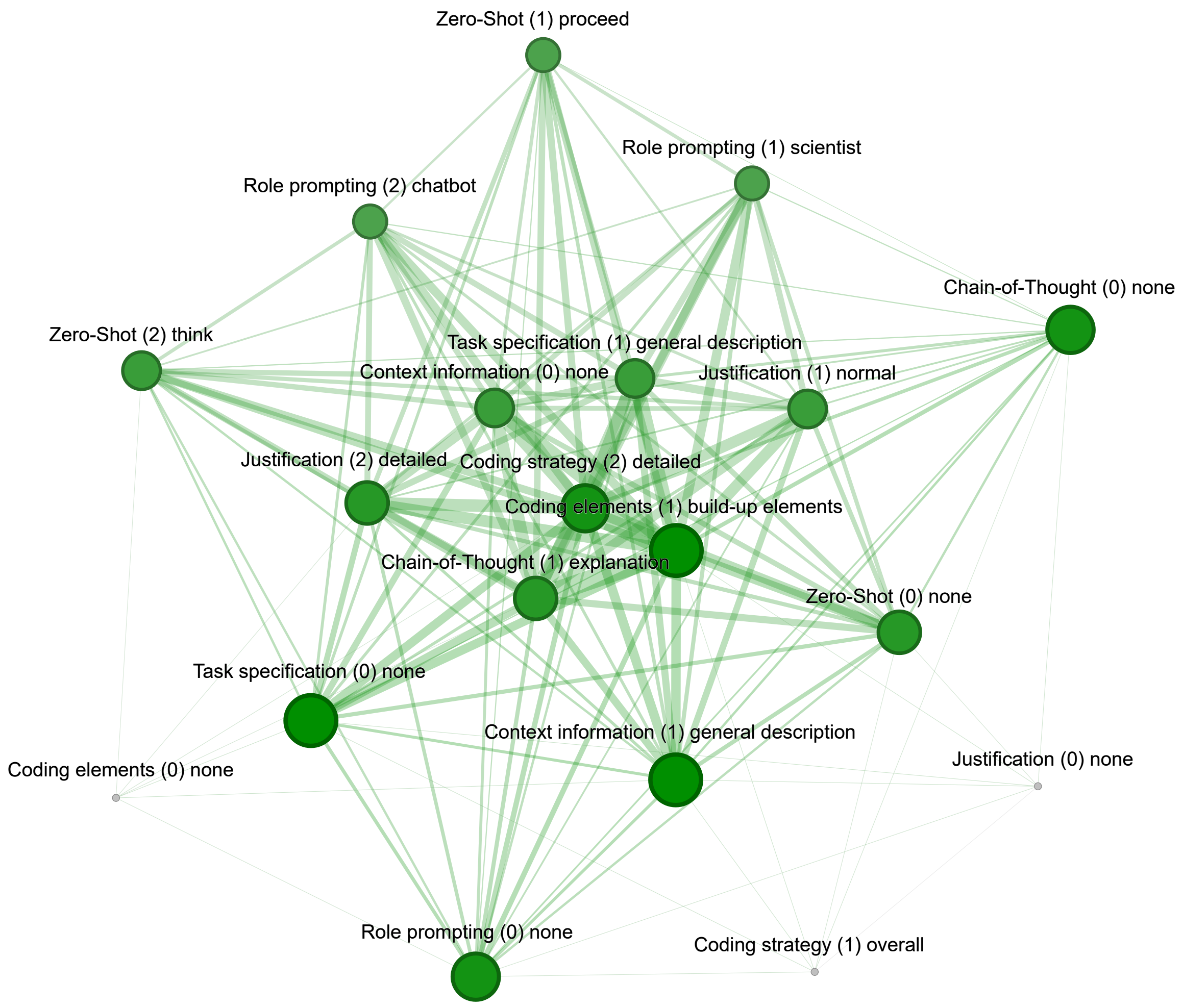

**Figure A2.** Network of prompting strategies

*Note*. The network models connections between prompting strategies (nodes) and their frequency of co-occurrence (edges) regarding the dataset of acceptable reliability. Size and color of nodes refer to the number of connections (degrees), color of edges to the number of co-occurrences. We used the open-source tool *gephi* (Bastian et al., 2009) and *ForceAtlas2* as the layout algorithm.

**Table A3**. Node centrality measured by strength

| Prompting strategy | Node strength |
|---|---|
| Coding elements (1) build-up elements | 280 |
| Coding strategy (2) detailed | 280 |
| Chain-of-Thought (1) explanation | 238 |
| Task specification (1) general description | 175 |
| Context information (0) none | 168 |
| Justification (2) detailed | 147 |
| Justification (1) normal | 133 |
| Context information (1) general description | 119 |
| Task specification (0) none | 112 |
| Zero-Shot (0) none | 112 |
| Role prompting (1) scientist | 112 |
| Role prompting (2) chatbot | 105 |
| Zero-Shot (2) think | 91 |
| Zero-Shot (1) proceed | 84 |
| Role prompting (0) none | 70 |
| Chain-of-Thought (0) none | 49 |
| Justification (0) none | 7 |
| Coding elements (0) none | 7 |
| Coding strategy (1) overall | 7 |

**Table A4.** Occurrences of prompting strategies in common "best" prompts between variables

| | Prompting strategy | | | | | | | | | α | |
|---|---|---|---|---|---|---|---|---|---|---|---|
| | RP | CI | TS | CS | CE | ZSCT | CT | J | Mean | $v_{climate}$ | $v_{movement}$ |
| Value | 2 | 0 | 0 | 2 | 1 | 0 | 1 | 2 | .73 | .71 | .74 |
| | 1 | 1 | 1 | 2 | 1 | 1 | 1 | 1 | .72 | .72 | .72 |
| | 2 | 0 | 0 | 2 | 1 | 2 | 1 | 2 | .72 | .70 | .74 |
| | 1 | 0 | 1 | 2 | 1 | 2 | 1 | 1 | .71 | .71 | .70 |
| | 1 | 1 | 1 | 2 | 1 | 0 | 1 | 1 | .70 | .70 | .70 |
| | 2 | 0 | 1 | 2 | 1 | 2 | 1 | 2 | .69 | .68 | .70 |
| | 2 | 1 | 1 | 2 | 1 | 1 | 1 | 2 | .69 | .67 | .70 |
| | 1 | 0 | 0 | 2 | 1 | 1 | 1 | 2 | .68 | .69 | .68 |

*Note*. RP = role prompting; CI = context information; TS = task specification; CS = coding strategy; CE = coding elements; ZSCT = Zero-Shot; CT = Chain-of-Thought; J = justification.

**Table A5.** Common best prompt (translated) for $v_{climate}$ and $v_{movement}$

| Variable | Prompt |
|---|---|
| $v_{climate}$ | You are a German chatbot. Decide for each reader comment whether the following characteristics are present. 1) Explanation of climate change (e.g., long-term changes in temperatures and weather patterns) (explanation_climate_change. 2) Causes of climate change (e.g., pollution, greenhouse gas emissions, deforestation, global warming) (causes_climate_change). 3) Effects and consequences of climate change (e.g., wildfire, extreme weather events, temperature increases, droughts, floods) (signs_climate_change). 4) Individual, social, or political actions to prevent or mitigate climate change and its consequences (e.g., not eating meat, not flying, increasing renewable energy, climate protection, environmental protection) (measures_climate_change). 5) Evaluation of climate change (e.g. relativization of the importance of climate change compared to other topics) (evaluation_climate_chance). 1) Read the category description and the reader's comment. 2) Based on the category description, think about whether the reader's comment is about climate change. 3) Type the appropriate answer. For each characteristic, indicate whether it is present in the comment (true) or not (false). Justify your decision in detail. Answer using the following pattern in json format: {'reason': string, 'explanation_climate_change': boolean, 'causes_climate_change': boolean, 'signs_climate_change': boolean, |
| | 'measures_climate_change': boolean, 'evaluation_climate_change': boolean}. Here is the comment you should analyze: |
| $v_{movement}$ | You are a German chatbot. Decide for each reader comment whether the following characteristics are present. 1) Mention of the movement (e.g. Fridays for Future, Last Generation or other names of the movements such as Climate Stickers, Climate RAF or Last Generation) (naming_movement). 2) Goals of the movement that are specifically related to the movement (e.g. adherence to the Paris Climate Agreement, concrete demands for climate protection, concrete demands to politicians, explicit accusations against politicians) (goals_movement). 3) Thematization of actions of the movements (e.g. demonstrations, climate strikes, school strikes, vandalism, road blockades) (thematization_action). 4) Evaluation of the movement, its goals and/or its actions (e.g. criticism and/or praise of the movement and/or its actions, description of the movement's goals as useful and/or useless) (evaluation_movement). 1) Read the category description and the reader's comment. 2) Based on the category description, think about whether the reader's comment is about Fridays for Future (FFF) and/or Last Generation (LG). 3) Type the appropriate answer. For each characteristic, indicate whether it is present in the comment (true) or not (false). Justify your decision in detail. Answer using the following pattern in json format: {'justification': string, 'naming_movement': boolean, 'goals_movement': boolean, 'thematization_action': boolean, 'evaluation_movement': boolean}. Here is the comment you should analyze: |

**Table A6.** Key evaluation metrics for the common best prompt for $v_{climate}$ and $v_{movement}$

| Evaluation metric | $v_{climate}$ | $v_{movement}$ |
|---|---|---|
| Accuracy | .88 | .87 |
| F1 Score | .79 | .87 |
| α | .71 | .74 |
| Precision | .79 | .92 |
| Recall | .79 | .83 |

**Table A7.** Prompts used in the Main study

| Task | Prompt |
|---|---|
| PImPo Migration detection | You are an English chatbot. Decide for the manifesto sentence whether the following characteristic is present.<br><br>1) Migration (migration): This characteristic is present if the sentence addresses immigration and/or immigrant integration. According to the codebook, this includes the regulation of entry, permanent residency of |

non-citizens, and the life of migrants and their descendants already residing in the country.

Indicators for 1:
- Entry & Border Management: Regulation of legal or unauthorized entry, border controls, visa regulations, Frontex, and freedom of movement within the EU.
- Groups: Asylum-seekers, refugees, labor migrants, exchange students, and family reunification.
- Integration & Membership: Naturalization, citizenship tests, language/orientation courses, and the social, political, or economic inclusion of migrants.
- Rights & Culture: Anti-discrimination/anti-racism specifically for migrants, religious rights (e.g., mosques), multiculturalism, and national identity in the context of migration.
- Causes & Removal: Addressing the causes of migration/displacement in origin countries, and the deportation or expulsion of non-citizens for any reason.
- Narratives: General approval or disapproval of a multicultural society or being a "nation of immigrants."

0) Neither (neither): This characteristic is present if the sentence does NOT relate to migration or integration.

Exclusions for 0:
- General Policy: Policy addressing the general population, "all citizens," or "all children" (e.g., general healthcare, education, or taxes) without specific focus on migrants.
- Autochthonous/Native Groups: Issues concerning indigenous populations, native tribes, or traditional ethnic minorities that are not migrants.
- Other Vulnerable Groups: Support for veterans, seniors, LGBTQ, or the disabled unless specifically linked to migration status.
- General Foreign Policy: International trade or military operations without a direct link to the movement of refugees.
- Emigration: Citizens of the country moving abroad.

1) Read the category descriptions and the sentence context.
2) Based on the category description, think about whether the TARGET SENTENCE belongs to category 1 or 0.
Decision Rule: Be conservative with general civics. If a sentence discusses "all people" or "our nation's values" without an explicit link to migrants or borders, choose 0.
3) Type the appropriate answer using the label 1 for migration and the label 0 for neither.

Justify your decision in detail.

| | |
|---|---|
| | Answer using the following pattern in json format: {{"reasoning": string, "decision": number}}.<br><br>Here is the context and the target sentence you should analyze:<br>Context Before: {context_before}<br>TARGET SENTENCE: "{target_sentence}"<br><br>Context After: {context_after} |
| PImPo Subtopic detection | You are an English chatbot. Decide for the migration-related manifesto sentence which of the following two categories is the primary focus.<br><br>1) Immigration (immigration): This category concerns the regulation of ADMISSION and legal entry into the country.<br>- Focus: The "Entry" dimension (who is allowed to come and stay).<br>- Indicators: Border management, visa regulations, asylum seeker/refugee status, recruitment of foreign labor, and family reunification (bringing people from abroad).<br>- Status Removal: Deportation or expulsion based purely on residence permits or unauthorized entry belongs here.<br>Examples: "Secure the borders," "New work visas for engineers," "Speed up asylum procedures," "Family reunification is a right."<br><br>2) Immigrant Integration (integration): This category concerns the conditions of RESIDENCY and the inclusion of people already living in the country.<br>- Focus: The "Membership" dimension (how people live and belong once they are inside).<br>- Indicators: Naturalization and citizenship (becoming a citizen), language courses, social/political participation, access to schools/welfare for residents, and anti-discrimination or anti-racism measures.<br>- Society: Multiculturalism, diversity, and religious freedoms (e.g., mosques).<br>- Conduct Removal: Deportation or removal of residence permits for long-term residents based on criminal behavior belongs here.<br>Examples: "Citizenship after five years," "Language courses should be mandatory," "Fight discrimination in the workplace," "Diversity has enriched us."<br><br>Decision Rule:<br>Ask: Is the sentence about "who gets to come in" (1) or "how we live together/become citizens" (2)?<br>- If the sentence focuses on the legal status of people living here (e.g. naturalization, rights of residents), choose 2.<br>- If the sentence focuses on the process of entry, visas, or border security, choose 1. |
| | 1) Read the category descriptions. 2) Analyze the TARGET SENTENCE. 3) Based on the focus (Entry vs. Membership), type the appropriate answer: 1 for immigration or 2 for integration.<br><br>Justify your decision in detail.<br><br>Answer using the following pattern in json format: {{"reasoning": string, "decision": number}}.<br><br>Here is the context and the target sentence you should analyze:<br>Context Before: {context_before}<br>TARGET SENTENCE: "{target_sentence}"<br>Context After: {context_after} |
| DAX-30 Tone Analysis | Du bist ein deutscher Chatbot.<br><br>Entscheide für den vorliegenden Textabschnitt aus einem Geschäftsbericht, welches der folgenden fünf Merkmale am ehesten zutrifft. Sie sind jeweils mit "---" getrennt und müssen in dieser hierarchischen Reihenfolge geprüft werden:<br><br>---<br><br>1. Kein konkreter Unternehmensbezug. Codiere hierfür 00.<br><br>- Ein Absatz gilt als ohne Unternehmensbezug, wenn der Inhalt sich nicht auf das spezifische Unternehmen, seine Einheiten oder Aktivitäten bezieht (z. B. allgemeine Marktsituation).<br>- Textbeispiel: "Das Börsenjahr 2014 war von Zuwächsen und zeitweilig großen Schwankungen geprägt."<br><br>---<br><br>2. Neutraler Absatz mit konkretem Unternehmensbezug. Codiere hierfür 01.<br><br>- Der Inhalt bezieht sich unmittelbar auf das Unternehmen (Name, "wir", "unser Unternehmen", eigene Zahlen), aber der Autor schildert sachlich-neutral ohne Wertung des dargestellten Sachverhaltes. Hierzu gehört auch die bloße numerische Darstellung von Veränderungen ohne wertende Einordnung. Insbesondere kleine Änderungen sind fast immer neutral.<br>- Textbeispiel: "Anstieg von 19,45 Euro auf 19,78 Euro"<br>- Textbeispiel: "Die außerplanmäßigen Abschreibungen betrugen für das abgelaufene Geschäftsjahr insgesamt 100 Mio EUR (Vorjahr: 137 Mio. EUR)."<br>- Textbeispiel: "Wir verfolgen die Aktienkursentwicklung auch im Vergleich zu unseren börsennotierten Wettbewerbern. Hierzu gehören vor allem die nordamerikanischen Düngemittelproduzenten PotashCorp und Mosaic, |

der russische Kaliproduzent Uralkali sowie der Salzhersteller Compass aus den USA."

Merke: Die explizite Relation zum Vorjahr (z. B. Vorjahreswerte in Klammern im Bezug auf eindeutig positive Unternehmensergebnisse oder Formulierungen wie „gegenüber dem Vorjahr“) gilt bereits als verbale Einordnung und ist gemäß Regel 3, 4 oder 5 zu codieren während reine (oft kleine) Zahlenänderungen neutral zu werten sind.

---

3. Positiver Absatz mit konkretem Unternehmensbezug. Codiere hierfür 11.

- Es liegt eine eindeutig positive Wertung des dargestellten Sachverhalts vor (Wachstum, Erfolg, Gewinn, Steigerung).
- Eine positive Wertung entsteht auch durch die verbale Einordnung eines Ergebnisses in Relation zum Vorjahr (z. B. Steigerung um 16 Prozent).
- Wird ein negativer Aspekt aufgegriffen, der das ebenfalls beschriebene Positive jedoch nur noch verstärkt (z. B. "Wir sind gewachsen, obwohl der Markt schwierig war"), wird der Absatz insgesamt als POSITIV (11) codiert. Im Mittelpunkt steht immer die Perspektive des Unternehmens, sein Handeln und seine Entwicklung.
- Textbeispiel: "Die Refinanzierungsaktivitäten des Jahres 2014 waren geprägt vom anhaltenden Wachstum des Volkswagen Konzerns."
- Textbeispiel: "Wir wollten das Ergebnis erneut steigern. Dies ist uns trotz der enttäuschenden Entwicklung in Europa gelungen: Wir sind gewachsen."
- Textbeispiel: "Insgesamt war der Markt sehr schwierig, aber wir konnten uns gut halten."
- Textbeispiel: "Der Konzernumsatz stieg währungsbereinigt um 16 Prozent auf 23,2 Milliarden Euro."

---

4. Negativer Absatz mit konkretem Unternehmensbezug. Codiere hierfür 21.

- Es liegt eine eindeutig negative Wertung des dargestellten Sachverhalts vor (Rückgang, Verlust, Enttäuschung, Verfehlen von Zielen, rückläufige Entwicklung).
- Eine negative Wertung entsteht auch durch die verbale Einordnung eines Ergebnisses in Relation zum Vorjahr (z. B. Rückgang um 16 Prozent).
- Wird ein positiver Aspekt aufgegriffen, der das ebenfalls beschriebene Negative jedoch nur noch verstärkt (z. B. "Obwohl die Bedingungen am Markt gut waren, konnten wir nicht abliefern"), wird der Absatz insgesamt als NEGATIV (21) codiert. Im Mittelpunkt steht immer die Perspektive des Unternehmens, sein Handeln und seine Entwicklung.
- Textbeispiel: "Infolgedessen beendete die Aktie der adidas AG das Jahr 2014 mit 57,62 Euro und lag um 38 % unter dem Vorjahreswert. Damit lag die Marktkapitalisierung des Konzerns am Jahresende bei 11,8 Mrd Euro (2013 19,4 Mrd Euro)."
- Textbeispiel: "Die Kurse der Stamm- und der Vorzugsaktie der Volkswagen AG haben sich 2014 bei hoher Volatilität insgesamt rückläufig entwickelt und blieben hinter der Entwicklung des Gesamtmarktes und der Automobilbranche zurück."
- Textbeispiel: "Im Vorjahr hatten wir noch Gewinne gemacht. In diesem Jahr haben wir diese Ziel leider verfehlt."
- Textbeispiel: "Infolgedessen beendete die Aktie der adidas AG das Jahr 2014 mit 57,62 Euro und lag um 38 % unter dem Vorjahreswert."

LOGISCHE RICHTUNG: Beachte die inhaltliche Bedeutung: Eine Steigerung von belastenden Faktoren (z. B. Kosten, Schulden, Steueraufwand, Rückstellungen) stellt faktisch eine Verschlechterung dar und ist als negativ (21) zu codieren.

---

5. Ambivalenter oder unklarer Absatz. Codiere hierfür 31.

- Innerhalb des Absatzes treten sowohl positive als auch negative Tonalitäten gleichwertig nebeneinander auf (Indikatoren: "trotz", "obwohl").
- Falls ein Absatz mit Unternehmensbezug absolut nicht eindeutig einer positiven oder negativen Ausprägung zugeordnet werden kann, wähle ebenfalls AMBIVALENT/UNKLAR (31).

- WICHTIG: Falls im Text sowohl positive als auch negative Informationen als eigenständige Fakten enthalten sind, ist zwingend 31 zu wählen. Es ist nicht möglich, eine Seite subjektiv als „wichtiger“ zu gewichten, um eine eindeutige Entscheidung für 11 oder 21 zu erzwingen. Beispiel: Die positiven Aspekte stehen im Vordergrund, der negative Teil ist nur eine Randnotiz oder stand am Anfang des Absatzes. Ein solcher Fall ist dennoch als 31 zu codieren.

---

Positive Wörter:

Allerbeste angehoben ankurbeln ansehnlich Anstieg Attraktivität ausgebaut ausgezeichnet bahnbrechend beachtenswert beachtlich bedeutend Belohnung Benefiz Bereicherung Besonders Beste/r Bonus Boom Chancen genutzt eindrucksvoll einwandfrei Engagement Erfolg

Erhöhung ertragreich exzellent Flexibilität Fortschritt Freude fruchtbar gelungen Gewinn gesteigert herausragend hervorragend Highlight Höchst(...)?(Ausnahme: Höchststand, außer Wertung steht davor, beispielsweise historischer) höher honorieren innovativ Kompetenz leistungsstark lobenswert lohnend lukrativ Maximum namhaft nutzbringend optimal (ein) Plus von.. positiv Profit profitieren Qualität renommiert Rendite – nur wenn Wertung davor steht rentabel sagenhaft Siegeszug Sicherheit Stabilität Steigerung Stolz Stärke/ am stärksten tadellos Triumph übertreffen überdurchschnittlich verbessern / Verbesserung verdient verdoppelt Vergrößerung Vermögen – nur wenn Wertung davor steht Vorteil Wachstum wertvoll Wertsteigerung Wohlstand Zusammenarbeit zurück / Rückgang Zuwachs

Negative Wörter:

Abbau Abbruch Abhängigkeit Ablehnung Abnahme Abschaffen Absenkung Abstieg Absturz anfällig Arbeitslose aufgeben Auflösung aussichtslos beenden Bedrohung Befürchtung Belastung besorgniserregend beunruhigend chaotisch Defizit drastisch düster einbrechen Einschränkung einstürzen Entlassung Enttäuschung erschreckend erschütternd Fehler fragwürdig gering Gefahr grausam heikel ineffizient instabil Kapitulation Katastrophe kippen kleiner Knappheit kompliziert Konflikt Konkurrenzkampf kontraproduktiv kostenintensiv kritisch labil leiden leider machtlos Mangel mindern mühsam nachlassen nachteilig negativ nicht erreicht niedrig nutzlos Pleite problematisch Risiko reduzieren Reinfall Ruin Rückgang rückläufig Schaden schlimm schockierend Schwierigkeit Schwäche sinken Sorge Stillstand unrentabel unsicher / verunsichert

Gehe Schritt für Schritt vor:

1. Analysiere, ob ein Unternehmensbezug vorliegt.
2. Prüfe auf neutrale Darstellung vs. Bewertung.
3. Prüfe auf Ambivalenz oder Unklarheit.
4. Wähle den passenden Code (00, 01, 11, 21 oder 31).
5. Prüfe noch einmal kritisch deine Entscheidung

Begründe deine Entscheidung kurz mit explizitem Bezug auf die genannten Regeln.

Antworte ausschließlich im folgenden JSON-Format:
{{"Begruendung": "...", "Entscheidung": "string_of_number"}}

Hier ist der Textabschnitt, den du analysieren sollst:
"{input_text}"

JSON-Antwort: